\documentclass[letterpaper, 10 pt, conference]{ieeeconf}  

\IEEEoverridecommandlockouts 

\usepackage{graphicx} 
\graphicspath{{figs/}}
\usepackage[fleqn]{amsmath} 
\usepackage{amssymb}  
\usepackage[fleqn]{mathtools}

\usepackage{multicol}
\usepackage{multirow}
\usepackage{placeins}
\usepackage{diagbox}
\usepackage{todonotes}
\usepackage{adjustbox}
\usepackage{verbatim}
\usepackage{cite}
\usepackage{wrapfig}
\usepackage{booktabs}
\usepackage{url}
\usepackage[utf8x]{inputenc}

\usepackage{balance}

\usepackage{algorithmic}

\usepackage[noline,linesnumbered]{algorithm2e}
\SetArgSty{textnormal} 

\newcommand{\cvec}[1]{\boldsymbol{\mathrm{#1}}}
\DeclareMathAlphabet{\mathpzc}{OT1}{pzc}{m}{it}
\DeclareMathAlphabet{\mathcall}{OMS}{cmsy}{m}{n}

\usepackage{amsmath}
\usepackage{amssymb}
\usepackage{dblfloatfix}
\usepackage{tikz}
\usepackage{subcaption}
\usepackage{textcomp}
\usepackage{url}
\usepackage[%
  xindy,
  section = section,
  nonumberlist,
  sanitize={symbol=false},
  shortcuts,
  acronym,
  nomain,
  nowarn
]{glossaries}                           

\newacronym{COPOS}{COPOS}{compatible policy search}
\newacronym{DDP}{DDP}{differential dynamic programming}
\newacronym{DQN}{DQN}{Deep $Q$-Network}
\newacronym{HSVI}{HSVI}{Heuristic Search Value Iteration}
\newacronym{GRL}{PO-GRL}{partially observable guided reinforcement learning}
\newacronym{LSTM}{LSTM}{Long Short-Term Memory}
\newacronym[\glslongpluralkey={Markov decision processes}]{MDP}{MDP}{Markov decision process}
\newacronym{MPC}{MPC}{model predictive control}
\newacronym[\glslongpluralkey={partially observable Markov decision processes}]{POMDP}{POMDP}{partially observable Markov decision process}
\newacronym{PPO}{PPO}{Proximal Policy Optimization}
\newacronym{RL}{RL}{reinforcement learning}
\newacronym[\glslongpluralkey={recurrent neural networks}]{RNN}{RNN}{recurrent neural network}
\newacronym{SAC}{SAC}{Soft Actor-Critic}
\newacronym{TRPO}{TRPO}{Trust Region Policy Optimization}
\newglossary[syg]{symbol}{sys}{syo}{List of Symbols}
\newglossaryentry{acspace}{
  sort        = {space action},
  name        = {\ensuremath{\mathpzc{A}}},
  description = {set of all possible agent actions},
  type        = symbol
}
\newglossaryentry{sspace}{
  sort        = {space state},
  name        = {\ensuremath{\mathpzc{S}}},
  description = {set of all possible environment states},
  type        = symbol
}
\newglossaryentry{obspace}{
  sort        = {space observation},
  name        = {\ensuremath{\mathpzc{O}}},
  description = {set of all possible observations},
  type        = symbol
}
\newglossaryentry{ac}{
  sort        = {vec action},
  name        = {\ensuremath{\cvec{a}}},
  description = {action taken by the agent},
  type        = symbol
}
\newglossaryentry{s}{
  sort        = {vec state},
  name        = {\ensuremath{\cvec{s}}},
  description = {environment's state},
  type        = symbol
}
\newglossaryentry{ob}{
  sort        = {vec observation},
  name        = {\ensuremath{\cvec{o}}},
  description = {observation},
  type        = symbol
}
\newglossaryentry{h}{
  sort        = {vec history},
  name        = {\ensuremath{\cvec{h}}},
  description = {history of observations},
  type        = symbol
}

\title{Reinforcement Learning using Guided Observability}

\author{Stephan Weigand$^{1}$, Pascal Klink$^{1}$, Jan Peters$^{1,2}$, Joni Pajarinen$^{1,3}$
\thanks{This work was supported by ERC StG \#640554 (SKILLS4ROBOTS) and DFG project PA 3179/1-1 (ROBOLEAP)}%
\thanks{$^{1}$Intelligent Autonomous Systems, TU Darmstadt, Germany}
\thanks{$^{2}$MPI for Intelligent Systems, Tuebingen, Germany}
\thanks{$^{3}$Department of Electrical Engineering and Automation, Aalto University, Finland}
\thanks{{\tt\small \{stephan.weigand,klink,peters,pajarinen\} @ ias.tu-darmstadt.de}}
}%

\begin{document}

\maketitle
\thispagestyle{empty}
\pagestyle{empty}

\begin{abstract}
  Due to recent breakthroughs, \gls{RL} has demonstrated impressive
  performance in challenging sequential decision-making
  problems. However, an open question is how to make \gls{RL} cope
  with partial observability which is prevalent in many real-world
  problems. Contrary to contemporary \gls{RL} approaches, which focus
  mostly on improved memory representations or strong assumptions
  about the type of partial observability, we propose a simple but
  efficient approach that can be applied together with a wide variety
  of \gls{RL} methods. Our main insight is that smoothly transitioning
  from full observability to partial observability during the training
  process yields a high performance policy. The approach,
  called \gls{GRL}, allows to utilize full state information during
  policy optimization without compromising the optimality of the final
  policy. A comprehensive evaluation in discrete \gls{POMDP} benchmark
  problems and continuous partially observable MuJoCo and OpenAI gym
  tasks shows that \gls{GRL} improves performance. Finally, we
  demonstrate \gls{GRL} in the ball-in-the-cup task on a real Barrett WAM
  robot under partial observability.
\end{abstract}

\section{Introduction}\label{sec:intro}

\Gls{RL} is a special field of machine learning where an agent, e.g.\ a
robot, tries to learn behavior through interaction with an
environment. The robot does that by maximizing a reward signal that he
receives after each interaction. By making use of the powerful
function approximation properties of neural networks, deep \gls{RL}
algorithms can approximate optimal policies and/or value
functions. Deep \gls{RL} has achieved high attraction in recent years
due to breakthroughs in simulated
domains~\cite{silver2017mastering,vinyals2019grandmaster}.
In robotic tasks,
deep \gls{RL} has been able to learn control policies directly from
camera inputs \cite{levine2016end,levine2018learning}.

However, \emph{an open question is how to make \gls{RL} work well
  under partial observability}. In many real-world scenarios, it is
  not possible to perceive the exact state of the
  environment. An example scenario is driving an autonomous car in bad
  weather conditions where not all sensor data is available or only
  poor quality data is available. The common formal definition for
  such problems is a \gls{POMDP} \cite{kaelbling1998planning}.
For optimal decision making, \glspl{POMDP} require taking the complete
event history into account making policy optimization challenging.

\begin{figure}[t]
  \centering
  \vspace{0.5em} \hspace{2em}\includegraphics[width=0.85\columnwidth]{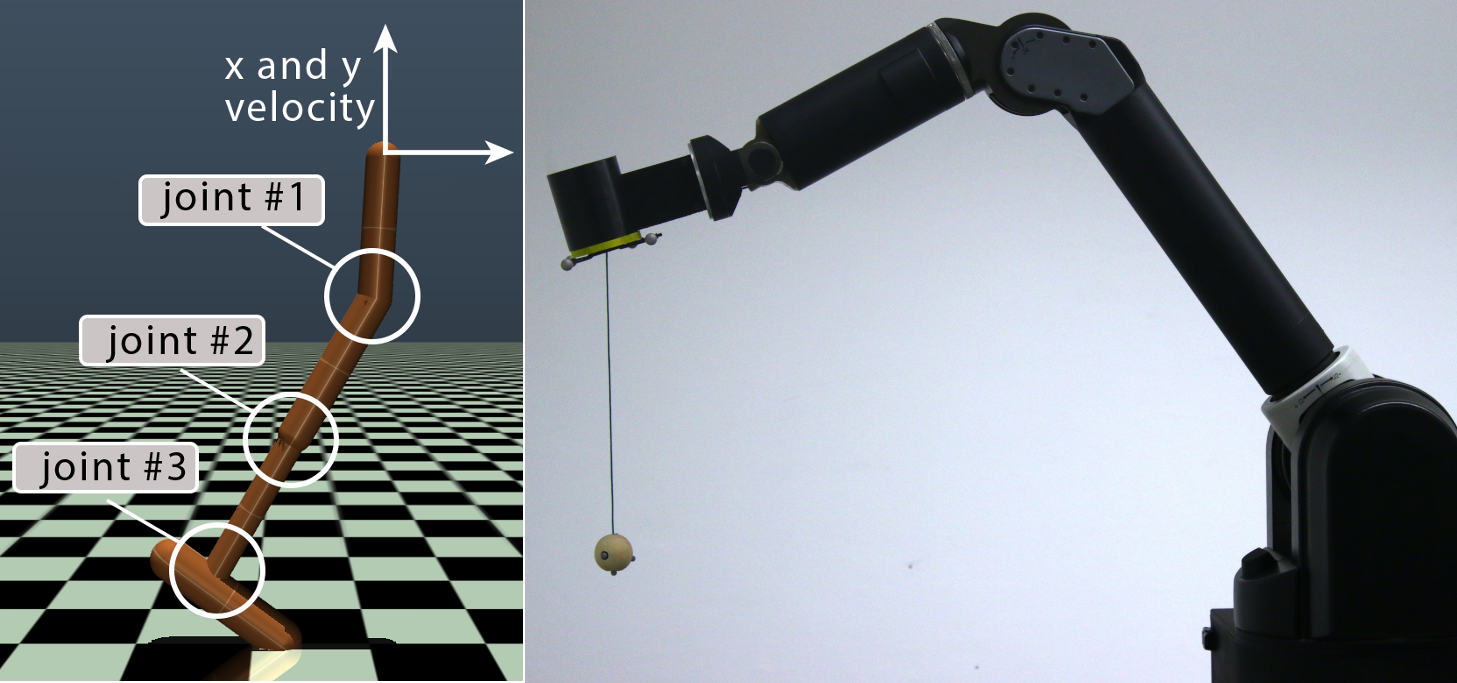}\\
  \vspace{0.5em} \includegraphics[width=0.95\columnwidth]{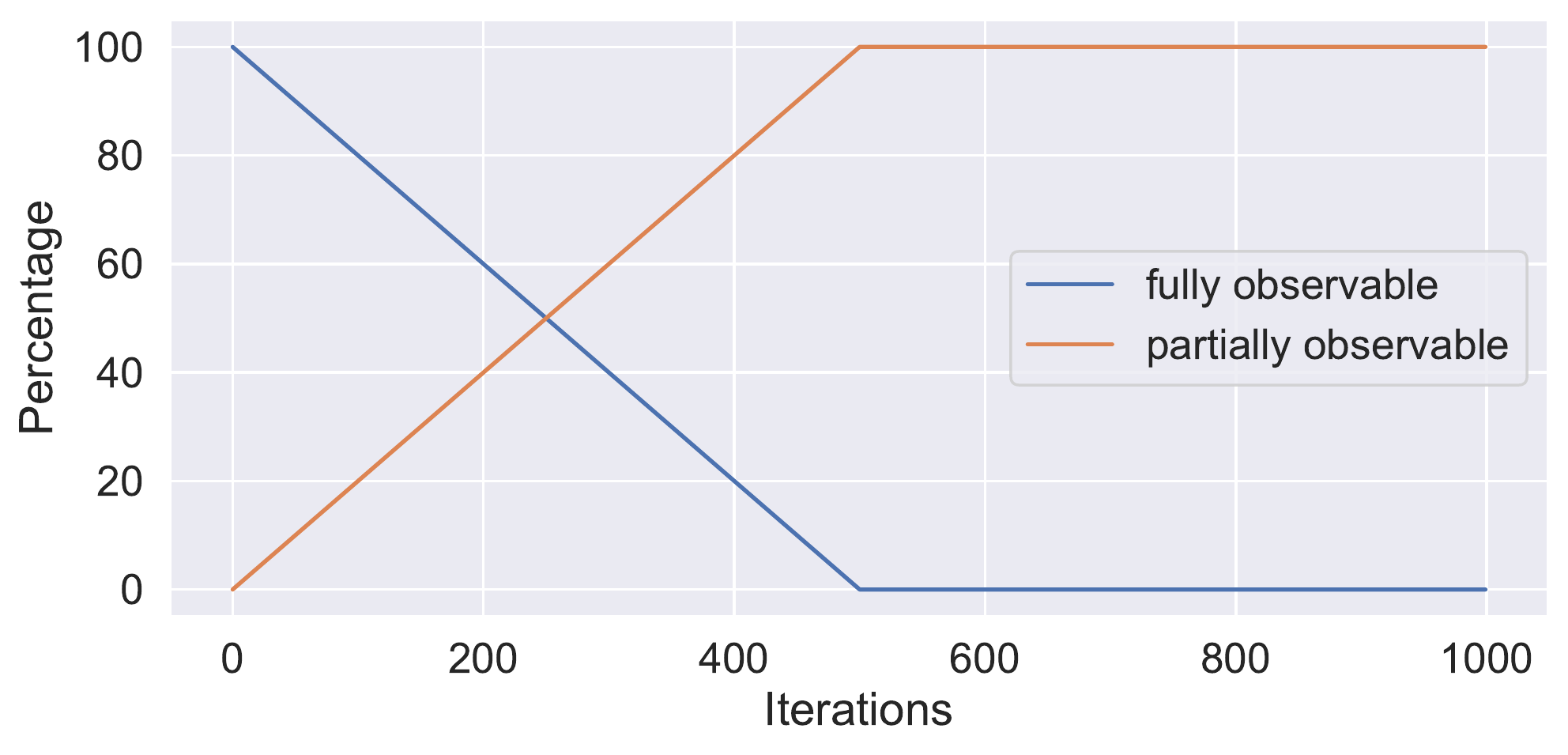}
\caption{We propose a reinforcement learning approach called \gls{GRL} to
         optimize robot behavior under partial
  observability. \textbf{(Bottom)} \gls{GRL} transitions from full
  observability to partial observability during optimization allowing
  to utilize full state information but ending
  up with a policy assuming solely partial observability. We
  evaluate \gls{GRL} in both discrete and continuous simulation tasks
  and on a real robot. \textbf{(Top Left)} An example of a continuous
  simulation task where the robot can not observe the highlighted joints
  and x and y velocity. \textbf{(Top Right)} Barrett WAM performing the
  ball-in-the-cup task with a partially observed ball.}
  \vspace{-1em} \label{fig:teaser}
\end{figure}

To make policy optimization more efficient we utilize full state
information during training. In many robotic applications, training
can be done in simulation with access to full state information,
and, the optimized policy transferred to the real system. Even without
(exact) simulation, physical systems can be set up to provide more
information during training compared to running the final
optimized policy. For example, in robotic multi-object manipulation
with visual sensory input objects are partially
observable \cite{pajarinen2017robotic}. However, the objects could be
marked with markers during training to yield full state
information. As another example, for hydraulic machines without
proprioception one could attach markers to robot links for full
state information during training. Similarly, mobile robots that rely
on visual input for localization \cite{radwan2018vlocnet++} could be
trained by using preconfigured measuring posts.

To take a step closer to solving \glspl{POMDP} we propose a simple but
efficient approach that utilizes full state information during
optimization. Classical work on \gls{RL} under partial
observability \cite{bakker2002reinforcement, wierstra2007solving,
hausknecht2015deep, heess2015memory} focuses on investigating
different memory representations such as \glspl{RNN} but does not
consider how to make the learning process more efficient. Other
methods are either application specific \cite{zhang2016learningmpc},
utilize models \cite{levine2013guided, levine2014learningcomplex}, or
make the policy assume information which is not available during
online operation~\cite{huttenrauch2017guided,pinto2018}. We focus on
model-free \gls{RL} that enables us to optimize policies directly
through system interactions without learning a model. We propose a
novel approach, called \gls{GRL}, which guides \gls{RL} algorithms
with additional full state information during the training phase. Our
approach, illustrated in Fig.~\ref{fig:teaser}, is based on mixing
full and partial state information while we gradually decrease the
amount of full state information during training ending up with a
policy compatible with partial observations. Our approach is able to
efficiently learn behavior for fully observable parts without
sacrificing performance in the partially observable parts of the
problem.

\section{Related Work}\label{sec:related-work}
The most naive strategy for dealing with partial observability is to
ignore it. \cite{kaelbling1996reinforcement} treats observations as if
they were states of the environment and \gls{MDP} methods are applied
to learn potentially sub-optimal behavior. So-called memory-based
approaches use finite memory representations to learn based on
previous interactions with the environment. \gls{LSTM} has been
studied widely as a memory
representation~\cite{wierstra2007solving,wierstra2007policy,hausknecht2015deep,heess2015memory,bakker2002reinforcement}. Even
with memory it is challenging to learn optimal
behavior. \cite{futoma2018prediction} proposes a prediction
constrained training for \glspl{POMDP} which allows effective model
learning even in the settings with misspecified models, as it can
ignore observation patterns that would distract classic two-stage
training. All these methods focus on memory representation rather than
on the optimization of the policy itself. As an alternative for
training an \gls{RNN} to summarize the past, \cite{igl2018deep} learns
a generative model of the environment and perform inference in this
model to effectively aggregate the available information.

Guided \gls{RL} approaches support \gls{RL} algorithms during the
learning process to increase their performance at test time. Guided
policy search interleaves model-based trajectory optimization and deep
learning for trajectory based
tasks \cite{levine2013guided,levine2014learningcomplex,levine1501learning,levine2016end,montgomery2016guided}. Guided
policy search has been extended with continuous-valued internal
memory \cite{zhang2016learning} for partially observable
tasks. \cite{zhang2016learningmpc} present an algorithm which
combines \gls{MPC} with guided policy search in the special domain of
quadcopters. The full state is known at training time when also MPC
is used, but unavailable at test time. While all these guided policy
search methods assume trajectory optimization with smooth
trajectories, we focus on a more general setting and also evaluate our
approach in benchmarks such as RockSample where trajectory
optimization is not possible.

Regarding utilizing full state information when optimizing
general \gls{RL} policies under partial observability,
\cite{huttenrauch2017guided} presents a guided actor-critic
approach for learning policies for a set of cooperative homogeneous
agents. While each actor learns a decentralized control policy
operation on (partially observable) locally sensed information, the
critic has access to the full system state. This means that the policy
is optimized to assume full state information at each future time
step. Such a policy cannot optimally consider information gathering
under partial
observability \cite{kaelbling1996reinforcement}. Similarly
to \cite{huttenrauch2017guided}, \cite{pinto2018} proposes to use a
guided actor-critic approach where the actor performs actions based on
image input while the critic has access to full state information. Due
to the critic being trained based on full state information the policy
can be, similarly to \cite{huttenrauch2017guided},
sub-optimal. Contrary to \cite{huttenrauch2017guided,pinto2018} our
optimization process starts by utilizing full state information but
\emph{decreases} the amount of full state information during optimization and
ends up with a policy that has been adapted to the partially
observable environment.

Finally, for completeness we mention that classic \gls{POMDP} methods
\cite{shani2013survey} assume a known model and
typically discrete actions and observations. In contrast, we focus on
model-free \gls{RL} that can handle continuous actions and
observations and directly learns from a reward signal.

\section{Preliminaries}\label{sec:background}
This section discusses background information needed to understand
our \gls{GRL} approach. First, we will describe the mathematical
framework for \glspl{POMDP}. Following standard
notation \cite{kaelbling1998planning, wiering2012reinforcement},
a \gls{POMDP} is a tuple $<\gls{sspace}, \gls{acspace}, \gls{obspace},
P, \mathcal{O}, r>$ consisting of states $\gls{s} \in \gls{sspace}$,
actions $\gls{ac} \in \gls{acspace}$, and observations
$\gls{ob} \in \gls{obspace}$. $P(\gls{s}' | \gls{s}, \gls{ac})$ is the
probability (for continuous values a probability density) of ending in
state $\gls{s}'$ when starting in state $\gls{s}$ and executing action
$\gls{ac}$. $\mathcal{O}(\gls{ob} | \gls{s}',\gls{ac})$ is the
probability of making observation $\gls{ob}$ given that the agent took
action $\gls{ac}$ and landed in state
$\gls{s}'$. $r(\gls{s}, \gls{ac})$ is the reward function.

As input for the underlaying algorithms, we maintain a history \gls{h}
containing the last $H$ observations (and actions), where the horizon
$H$ is an a-priori defined parameter. For further details see
section \ref{sec:experimental-setup}. The full history is a sufficient
statistic for optimal decision making in \glspl{POMDP} and a finite
history an approximation \cite{kaelbling1996reinforcement}. The goal
in \gls{RL} for \glspl{POMDP} is to find a policy $\pi(\gls{ac}_t
| \gls{h}_t)$ that maximizes the expected reward
$\mathbb{E}_{\gls{h}_{0}, \gls{ac}_0, ...}
[ \sum_{t=0}^{\infty} \gamma^t r(\gls{h}_t, \gls{ac}_t)]$. In
the continuous case, the expected reward can be
defined as \cite{schulman2015trust}
\begin{equation*}
J(\pi) = \iint p_{\pi}(\gls{h}) \pi(\gls{ac} | \gls{h}) Q^{\pi}(\gls{h}, \gls{ac}) d\gls{h}d\gls{ac},
\end{equation*}
where $p_{\pi}(\gls{h})$ denotes a (discounted) observation history
distribution induced by policy $\pi$ and
\begin{equation*}
\begin{split}
Q^{\pi}(\gls{h}_t, \gls{ac}_t) &= \mathbb{E}_{\gls{h}_{t+1}, \gls{ac}_{t+1}, ...} \Bigg[ \sum_{t=0}^{\infty} \gamma^t r(\gls{h}_t, \gls{ac}_t)\Bigg],\\
V^{\pi}(\gls{h}_t) &= \mathbb{E}_{\gls{ac}_{t}, \gls{h}_{t+1}, ...} \Bigg[ \sum_{t=0}^{\infty} \gamma^t r(\gls{h}_t, \gls{ac}_t)\Bigg],\\
A^{\pi}(\gls{h}_t, \gls{ac}_t) &= Q^{\pi}(\gls{h}_t, \gls{ac}_t) - V^{\pi}(\gls{h}_t)
\end{split}
\end{equation*}
denote the state-action (history-action in the \gls{POMDP} case) value
function $Q^{\pi}(\gls{h}_t, \gls{ac}_t)$, the value function
$V^{\pi}(\gls{h}_t)$ and the advantage function
$A^{\pi}(\gls{h}_t, \gls{ac}_t)$. For the discrete case, integrals can
be replaced by sums.

\section{Reinforcement Learning using Guided Observability}
This section presents our \gls{GRL} approach. During
training \gls{GRL} mixes both full and partial information until a
defined point (for example, half of training iterations) decreasing
linearly the amount of full information. For the remaining training
iterations (for example, final half of training iterations) \gls{GRL}
uses only partial information resulting in a policy fully compatible
with partial observations. We demonstrate the example usage with the
policy search algorithm \gls{COPOS} and the actor-critic algorithm
\gls{SAC}. \emph{\gls{GRL} has the same convergence guarantees as applying
the underlaying \gls{RL} algorithm directly on partial observations}
due to the full conditioning on partial observations.

We assume having access to both full and partial observations during
training and only to partial observations at test time. Partial
observability at test time can mean that parts of the full state
information are missing or noisy, or that the complete observations
are noisy.
Further, we assume training on a fixed number of samples. We also
assume that we know the dimensions of the observation vector that will
not be available, or noisy, during the test phase. For training
outside simulation, that is, in the real-world, it is often
straightforward to generate additional state information e.g.\ by
adding additional sensors or cameras in robotic tasks. By deactivating
parts of the observations at test time, we simulate practical
scenarios like broken sensors or omitting sensors to save costs. For
problems where the fully observable parts are not at all related to
the best solution, \gls{GRL} may not help.  However, when full state
information helps in solving some parts of the task our \gls{GRL}
approach may help underlying algorithms escape poor local optima
through additional state information. We will now describe our
approach based on samples $\tau$ consisting of tuples of
observations \gls{ob}, actions \gls{ac} and rewards r.

\paragraph{Algorithm Description}
Algorithm \ref{algo:grl} shows the \gls{GRL} approach for
batch-based \gls{RL} algorithms which we use as the guiding algorithm
template. The following procedure will be performed at the learning
phase: At the beginning of training, all samples $\tau$ contain fully
observable state information. The amount of full state information
decreases linearly until $N_{\textrm{MIX}}$ training iterations (see
Algorithm \ref{algo:grl}, Lines 4 and 5). From that point the training
will be continued with only partial information to optimize the final
policy for partial observability (see Algorithm \ref{algo:grl}, Line
7). Fig.~\ref{fig:teaser}(Bottom) visualizes this general
procedure. For simplicity, in the experiments we always mixed samples
until reaching half of the iterations in the learning phase. For
example, for 1000 training iterations the parameter $N_{\textrm{MIX}}$
will be set to 500. Interestingly, as shown in the experiments in
section~\ref{sec:experiments} the results are not sensitive to
$N_{\textrm{MIX}}$. In the future, depending on the problem, parameter
$N_{\textrm{MIX}}$ could be optionally fine-tuned. Now, we provide
further implementation details for two types of algorithms:
batch-based algorithms
like \gls{COPOS}, \gls{TRPO} \cite{schulman2015trust},
or \gls{PPO} \cite{schulman2017proximal} and algorithms using a replay
buffer like \gls{SAC}.

\paragraph{Implementation Details for COPOS}
Batch-based algorithms like \gls{COPOS} use a sample set
$\mathpzc{T}$, received from environment interactions in each
iteration.  Training in simulation, the environment returns always
fully observable samples (Algorithm \ref{algo:grl}, Line 3).  At
training iteration $i$, when $i < N_{\textrm{MIX}}$, we directly make
$|\mathpzc{T}| \cdot i / N_{\textrm{MIX}}$ samples partially
observable (Algorithm \ref{algo:grl}, Line 5) (when $i \ge
N_{\textrm{MIX}}$ we make all samples partially observable). For
partially observable samples, we overwrite partially observable
dimensions with zeros and set a flag. At test time the environment
directly return partially observable observations. We implemented the
\gls{COPOS} algorithm for discrete and continuous action spaces as well
as \gls{GRL} as a guided version of \gls{COPOS} using the OpenAI
baselines framework \cite{baselines}.
\begin{algorithm}
  \begin{algorithmic}[1]
    \STATE Initialize $\pi_0$, $N_{\textrm{MIX}}$
    \FOR{i = 0 \textbf{to} max iterations - 1}
      \STATE Generate sample set $\mathpzc{T}$ using $\pi_{i}$
      \IF{$i < N_{\textrm{MIX}}$}
        \STATE Make $|\mathpzc{T}| \cdot i / N_{\textrm{MIX}}$ samples partially observable
      \ELSE
        \STATE Make all samples partially observable
      \ENDIF
      \STATE Optimize and update $\pi_{i+1}$ using modified set $\mathpzc{T}$
    \ENDFOR
  \end{algorithmic}
  \caption{Partially Observable Guided Reinforcement Learning (PO-GRL)}
  \label{algo:grl}
\end{algorithm}

\paragraph{Implementation Details for SAC}
Some algorithms like \gls{SAC} have no explicit sampling phase to
collect individual sample sets for the optimization, instead they
perform the optimization based on a (fixed-length, rolling) replay
buffer where samples are collected after each environment
interaction. Training is performed on batches sampled from the replay
buffer. To apply \gls{GRL}, we can modify these batches to contain
$N_{\textrm{BATCH}} \cdot i / N_{\textrm{MIX}}$ partially observable
samples for training iterations $i < N_{\textrm{MIX}}$, where
$N_{\textrm{BATCH}}$ is the batch size, and for $i \ge
N_{\textrm{MIX}}$ we use only partially observable samples. However,
this modification of each batch can be computationally
heavy. Therefore, in the experiments, before $N_{\textrm{MIX}}$
training iterations, we make a sample partially observable already
before inserting it into the replay buffer. Samples are made partially
observable according to the probability $n_{\textrm{CURRENT}} /
n_{\textrm{MIX}}$, where $n_{\textrm{MIX}}$ is the total number of
samples obtained in $N_{\textrm{MIX}}$ training iterations and
$n_{\textrm{CURRENT}}$ is the index of the current sample. After
$N_{\textrm{MIX}}$ training iterations we use only partially
observable samples. Note, that for \gls{SAC}, the replay buffer
should not be too long, otherwise the optimization might still take
fully observable samples into account long after the last fully
observable sample was added to the buffer. In contrast, setting the
replay buffer length too short, the performance of \gls{SAC} will
decrease significantly. We modified the implementation of
the \gls{SAC} algorithm from the Stable Baselines
framework \cite{stable-baselines} using our \gls{GRL} approach.

\section{Experimental Results}\label{sec:experiments}
Our experimental evaluation was designed to answer the following
research questions:
\begin{enumerate}
\item How does our \gls{GRL} approach perform on \gls{POMDP} problems compared to other model-free \gls{RL} algorithms? Could our approach lead the policy search algorithm \gls{COPOS} and actor-critic algorithm \gls{SAC} to produce better results in contrast to learning only on partial state information? 
\item Is our approach useful when the observations contain symmetric noise? How does the application of our approach affect the results in comparison to training directly on noisy observations?
\end{enumerate}
To clarify the first question we analyzed the performance of our
approach and other \gls{RL} algorithms on different \gls{POMDP}
tasks. For answering the second question we designed a continuous
control task where all observations returned by the environment are
noisy. Also in this task, we compare our approach against training on
either noisy or the original state information.

\begin{table*}[hb]
  \small
  \centering
  \caption{Mean of the average return (average discounted return on
    discrete tasks) on nine fully observable tasks over the last 40
    episodes \textpm\ standard error over 50 random seeds.}
  \begin{tabular}{|lccccc|}
    \hline
    \textbf{Task} (Timesteps) & \textbf{COPOS} & \textbf{PPO} & \textbf{PPO-LSTM} &\textbf{SAC} & \textbf{TRPO} \\
    \hline
    \hline
    HalfCheetah-v2 (1M) &1629.53\textpm69.91 & 1303.73\textpm78.72 & 1460.23\textpm70.58 & \textbf{3830.97\textpm357.30} & 1009.25\textpm27.33 \\
    Hopper-v2 (1M) & 1899.34\textpm81.80 &1161.65\textpm75.75 & 1587.66\textpm58.28 & \textbf{2130.49\textpm61.89} & 1647.12\textpm38.42 \\
    InvertedDoublePendulum-v2 (1M) & \textbf{8112.95\textpm181.20} & 3367.50\textpm208.42 & 3889.17\textpm76.66 & 5164.28\textpm266.02 & 6589.08\textpm277.21 \\
    LunarLander-POMDP (5M) & 233.07\textpm1.29 & 26.06\textpm13.94 & 122.58\textpm39.07 & \textbf{240.06\textpm0.61} & 204.45\textpm3.52 \\
    Noisy-LunarLander (5M) & \textbf{255.61\textpm2.83} & 140.59\textpm11.15 & 53.12\textpm13.44 & - & 248.07\textpm3.26 \\
    Reacher-v2 (1M) & \textbf{-4.61\textpm0.04} & -6.37\textpm0.13 & -11.12\textpm0.09 & -6.86\textpm0.09 & -4.74\textpm0.05 \\
    RockSample(4,4) (3M) & \textbf{12.82\textpm0.28} & - & - & - & 9.03\textpm0.20 \\
    RockSample(5,7) (10M) & \textbf{13.13\textpm0.29} & - & - & - & 8.37\textpm0.16 \\
    Swimmer-v2 (1M) &101.03\textpm2.42 & 84.79\textpm3.26 & 72.49\textpm3.66 & 51.83\textpm2.42 & \textbf{114.95\textpm0.73} \\
    Walker2d-v2 (1M) & 1147.71\textpm89.68 & 871.69\textpm44.40 & 1049.95\textpm34.85 & \textbf{3740.61\textpm78.33} & 1331.77\textpm49.70 \\
    \hline
  \end{tabular}
  \label{tab:results-full}
\end{table*}

\begin{table*}[hb]
  \small
  \centering
  \caption{Mean of the average return (average discounted return on discrete tasks) on nine partially observable tasks over the last 40 episodes \textpm\ standard error over 50 random seeds.}
  \begin{tabular}{|lcccc|}
    \hline
    \textbf{Task} (Timesteps) & \textbf{COPOS} & \textbf{COPOS-guided} & \textbf{PPO} & \textbf{PPO-LSTM}\\
    \hline
    \hline
    HalfCheetah-v2 (1M) & 1490.79\textpm48.58 & 1491.02\textpm59.99 & 1314.10\textpm59.97 & 1157.11\textpm44.12 \\
    Hopper-v2 (1M) & 1411.96\textpm77.40 & \textbf{2219.61\textpm83.34} & 994.93 \textpm48.56 & 955.15\textpm33.31 \\
    InvertedDoublePendulum-v2 (1M) &5528.74\textpm297.16 & \textbf{7535.25\textpm241.52} & 2629.53\textpm133.94 & 935,93\textpm21.89 \\
    LunarLander-POMDP (5M) & 81.03\textpm20.29 & \textbf{154.16\textpm15.03} & -195.06\textpm17.99 & -272.95\textpm37.26 \\
    Noisy-LunarLander (10M) & 124.34\textpm1.76 & \textbf{132.74\textpm2.32} & -153.24\textpm12.89 & -117.54\textpm7.19 \\
    Reacher-v2 (1M) &-9.35\textpm0.05 & \textbf{-4.69\textpm0.05} & -9.66\textpm0.05 & -14.01\textpm0.06 \\
    RockSample(4,4) (3M) & 9.95\textpm0.28 & \textbf{11.83\textpm0.30} & - & - \\
    RockSample(5,7) (10M) & 8.07\textpm0.02 & \textbf{9.66\textpm0.21} & -& - \\
    Swimmer-v2 (1M) &60.95\textpm2.09 & \textbf{92.63\textpm2.96} & 38.04\textpm1.30 & 35.82\textpm0.45 \\
    Walker2d-v2 (1M) & 780.44\textpm46.69 & 993.29\textpm78.10 & 598.23\textpm18.06 & 699.56\textpm10.53 \\
    \hline
  \end{tabular}
  \begin{tabular}{|lccc|}
    \multicolumn{4}{c}{} \\
    \hline
    \textbf{Task} (Timesteps) & \textbf{SAC} & \textbf{SAC-guided} & \textbf{TRPO} \\
    \hline
    \hline
    HalfCheetah-v2 (1M) & 1536.54\textpm119.12 & \textbf{1835.62\textpm154.52} & 1049.87\textpm31.06 \\
    Hopper-v2 (1M) & 726.70\textpm37.76 & 858.00\textpm55.88 & 1038.55\textpm12.95 \\
    InvertedDoublePendulum-v2 (1M) & 6661.90\textpm161.32 & 5922.65\textpm252.54 & 5089.96\textpm190.03 \\
    LunarLander-POMDP (5M) & -70.13\textpm11.17 & - & -52.43\textpm15.86 \\
    Noisy-LunarLander (10M) & - & - & 123.00\textpm2.47 \\
    Reacher-v2 (1M) & -11.06\textpm0.20 & -11.58\textpm0.24 & -9.34\textpm0.06 \\
    RockSample(4,4) (3M) & - & - & 8.57\textpm0.25\\ 
    RockSample(5,7) (10M) & - & -  & 8.15\textpm0.00 \\
    Swimmer-v2 (1M) & 41.47\textpm0.77 & 42.25\textpm0.68 & 67.67\textpm1.19 \\
    Walker2d-v2 (1M) & 1892.36\textpm88.80 & \textbf{1911.40\textpm99.52} & 862.46\textpm27.01 \\
    \hline
  \end{tabular}
  \label{tab:results-partial}
\end{table*}

\subsection{Experimental Setup}\label{sec:experimental-setup}
For continuous action space experiments, we evaluate 
LunarLander-POMDP \cite{rongzhi}, Noisy-LunarLander and six tasks in
the MuJoCo \cite{todorov2012mujoco} physics simulator. The
LunarLander-POMDP environment is a modification that turns the OpenAI
Gym \cite{openaigym} environment LunarLanderContinuous-v2 into a
partially observable problem by adding a "blind" area between the
starting point on the top and the landing pad. Inside the the blind area
the lander agent cannot perceive any information of the
environment. Instead of using the original
observation output from the environment, we use a history of fixed
length as input for the \gls{RL} algorithms which is more suitable for solving
\glspl{POMDP}. The history includes the last $H$ observations
\gls{ob}, actions \gls{ac} and a flag $f$ that tells the algorithm
whether full or partial state information is available. We define the
used history for time step $t$ as $\boldsymbol{h}_t = \big( f_t,
\gls{ob}_t, \gls{ac}_{t-1}, ..., f_{t-H}, \gls{ob}_{t-H},
\gls{ac}_{t-H-1} \big)$. The history structure and the structure of
the neural networks remain always the same in all three cases (fully
observable, partially observable, guided) which enables us to focus on
the differences between partial and full observability. Based on the
LunarLanderContinuous-v2 environment we built another modification,
called Noisy-LunarLander, to answer the second research question posed
at the beginning of the section. We added Gaussian noise by randomly
drawing samples from a normal (Gaussian) distribution, represented by
the mean $\mu$ and the standard deviation $\sigma$. Additionally, we
focus on the two-dimensional tasks HalfCheetah-v2, Hopper-v2,
InvertedDoublePendulum-v2, Reacher-v2, Swimmer-v2 and Walker2d-v2 from
MuJoCo. In order to create partially observable environments, we
modified the MuJoCo tasks listed above by deactivating parts of the
observation. Specifically, we select a number of dimensions of the
observation vector and overwrite them with zeros. For the selection of
these dimensions for the individual MuJoCo tasks, we run some
preliminary tests and chose dimensions whose deactivation has a
noticeable impact on the overall
performance. Fig.~\ref{fig:teaser}(Top Left) visualizes these
dimensions for Hopper-v2.

As the discrete action task, we use RockSample, a well-known benchmark
problem for \gls{POMDP} algorithms which was first introduced in
\cite{smith2004heuristic}. We use the RockSample implementation from
gym-pomdp \cite{gympomdp} which are extensions of OpenAI Gym for
\glspl{POMDP} and built a wrapper which returns a history containing
the following information: The agent's observation for sampling or
checking rocks, the position of the agent in the grid, the previously
taken action and a flag that tells whether full or partial state
information is available. In the fully observable case, the true rock
values are additionally available. For our experiments, we use two
different sizes of the problem: RockSample(4,4) and
RockSample(5,7).

For each experimental run, we used one core on an
Intel\textsuperscript{\textregistered}
Xeon\textsuperscript{\textregistered} E5-2670 processor and a maximum
of 6 GB RAM. We denote the guided versions of the \gls{COPOS} and
\gls{SAC} algorithms "COPOS-guided" and "SAC-guided", respectively. We
compared against \gls{COPOS}, \gls{SAC}, \gls{TRPO}, and \gls{PPO},
where \gls{TRPO} and \gls{PPO} used the OpenAI baselines
\cite{baselines} implementation.  Additionally, to test whether a
different memory representation influences performance, we compared
against \gls{PPO} + \gls{LSTM}~\cite{hochreiter1997long} called
"PPO-LSTM" and implemented using Stable
Baselines \cite{stable-baselines}. In preliminary experiments we also
tried the OpenAI baselines implementation of \gls{PPO} + \gls{LSTM}
but achieved better results with Stable Baselines.  We ran all
experiments for 50 random seeds each.  For training, we used 1 million
time steps in the MuJoCo tasks, 5 million time steps in
LunarLander-POMDP, 10 million time steps in the Noisy-LunarLander
environment, 3 million time steps in the RockSample(4,4) environment
and 10 million time steps in the RockSample(5,7) environment.
The performance in the first 50\% timesteps for
COPOS-guided and SAC-guided is evaluated on a mixture of fully and
partially observable samples. The second half of iterations is
evaluated on only partially observable samples.

\subsection{Results}
\begin{figure}[ht]
  \centering
  \begin{subfigure}[b]{0.49\columnwidth}
    \centering
    \includegraphics[width=\columnwidth]{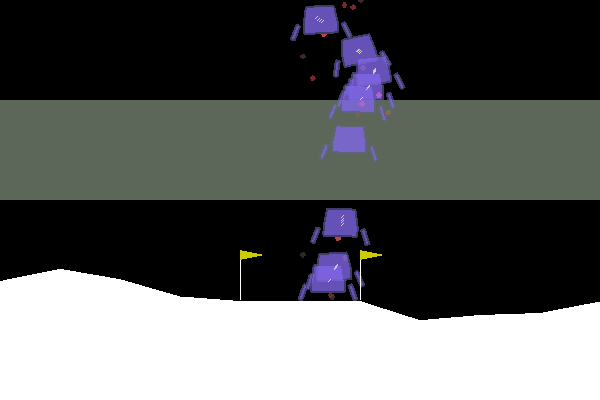}
    \caption{Lander is heavily breaking \\before blind area and falling
      through.}
    \label{fig:lunarlanderpomdp-viz-partial}
  \end{subfigure}
  \hfill
  \begin{subfigure}[b]{0.49\columnwidth}
    \centering
    \includegraphics[width=\columnwidth]{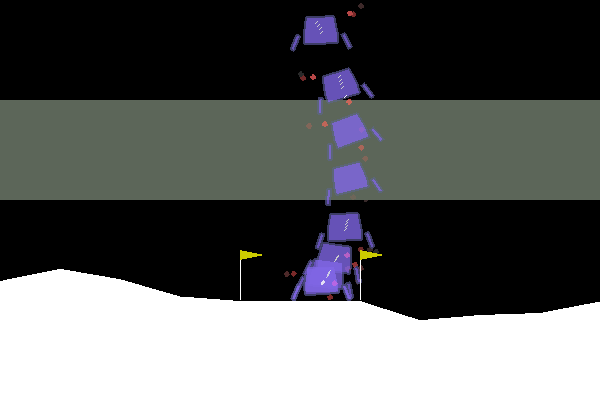}
    \caption{Speed of the lander is constant also inside blind area.\\}
    \label{fig:lunarlanderpomdp-viz-guided}
  \end{subfigure}
  \caption{Visualization of two trained policies in LunarLander-POMDP,
    where (a) is trained on partial information (COPOS) and (b) is
    trained with guided approach (COPOS-guided).}
\end{figure}

Table \ref{tab:results-partial} summarizes the results on all
partially observable tasks targeting the first research question. We
also ran all experiments in the fully observable case as baseline, see
Table \ref{tab:results-full}. In the LunarLander-POMDP environment, we
compared COPOS-guided with the algorithms \gls{COPOS}, \gls{PPO},
PPO-LSTM, \gls{SAC}, and \gls{TRPO} and rendered some runs of the LunarLander-POMDP (partially observable)
environment with the final learned policy for COPOS-guided (see Figure
\ref{fig:lunarlanderpomdp-viz-guided}) and \gls{COPOS} (see Figure
\ref{fig:lunarlanderpomdp-viz-partial}). There is a noticeable
difference in learned behavior: With the policy learned by
\gls{COPOS}, the lander moves slowly into the blind area, then it
falls fast through the blind area and after that slows down before
landing, while with the policy learned by COPOS-guided, the speed of
the lander is more constant also inside the blind area. Additionally,
with COPOS-guided the lander does not turn off the orientation engine
in the blind area as is done by COPOS in
Figure~\ref{fig:lunarlanderpomdp-viz-partial}.
We compared COPOS-guided and SAC-guided to \gls{COPOS}, \gls{PPO},
PPO-LSTM, \gls{SAC}, and \gls{TRPO}, on six different continuous
control tasks in the MuJoCo physics simulator.
Table~\ref{tab:results-partial} shows that our \gls{GRL} approach,
meaning either COPOS-guided or SAC-guided, improves the performance in
all partially observable MuJoCo tasks.

For evaluating our \gls{GRL} approach on environments with a discrete
action space, we compared COPOS-guided against \gls{COPOS} and
\gls{TRPO}. We were not able to test against \gls{SAC} because this
algorithm is only implemented for continuous action spaces. Table
\ref{tab:results-partial} demonstrates that COPOS-guided outperformed
the other algorithms in both problem sizes (for more details, see
Figure 1 in the supplement). Our approach supports the algorithm not
being stuck in poor local optima and therefore allows to learn a more
advanced policy. RockSample is a challenging task for model-free
approaches because there is no learned model of the environment
including the position of the rocks available. Therefore, approaches
which require an exact model already achieve better results on this
problem. For example, \gls{HSVI} \cite{smith2004heuristic} achieves a
reward of $18.0$ on RockSample(4,4) and a reward of $23.1$ on
RockSample(5,7).

To investigate our second research question, we compared COPOS-guided
against \gls{COPOS}, \gls{PPO}, PPO-LSTM and \gls{TRPO} in our
Noisy-LunarLander environment. For the Gaussian noise we use the
parameters $\mu = 0$ and $\sigma = 0.3$. Table
\ref{tab:results-partial} shows that the \gls{GRL} approach leads to a
higher performance than using \gls{COPOS}, \gls{PPO}, PPO-LSTM or
\gls{TRPO} directly on noisy data.

\begin{figure}[ht]
  \centering
  \includegraphics[width=0.9\columnwidth]{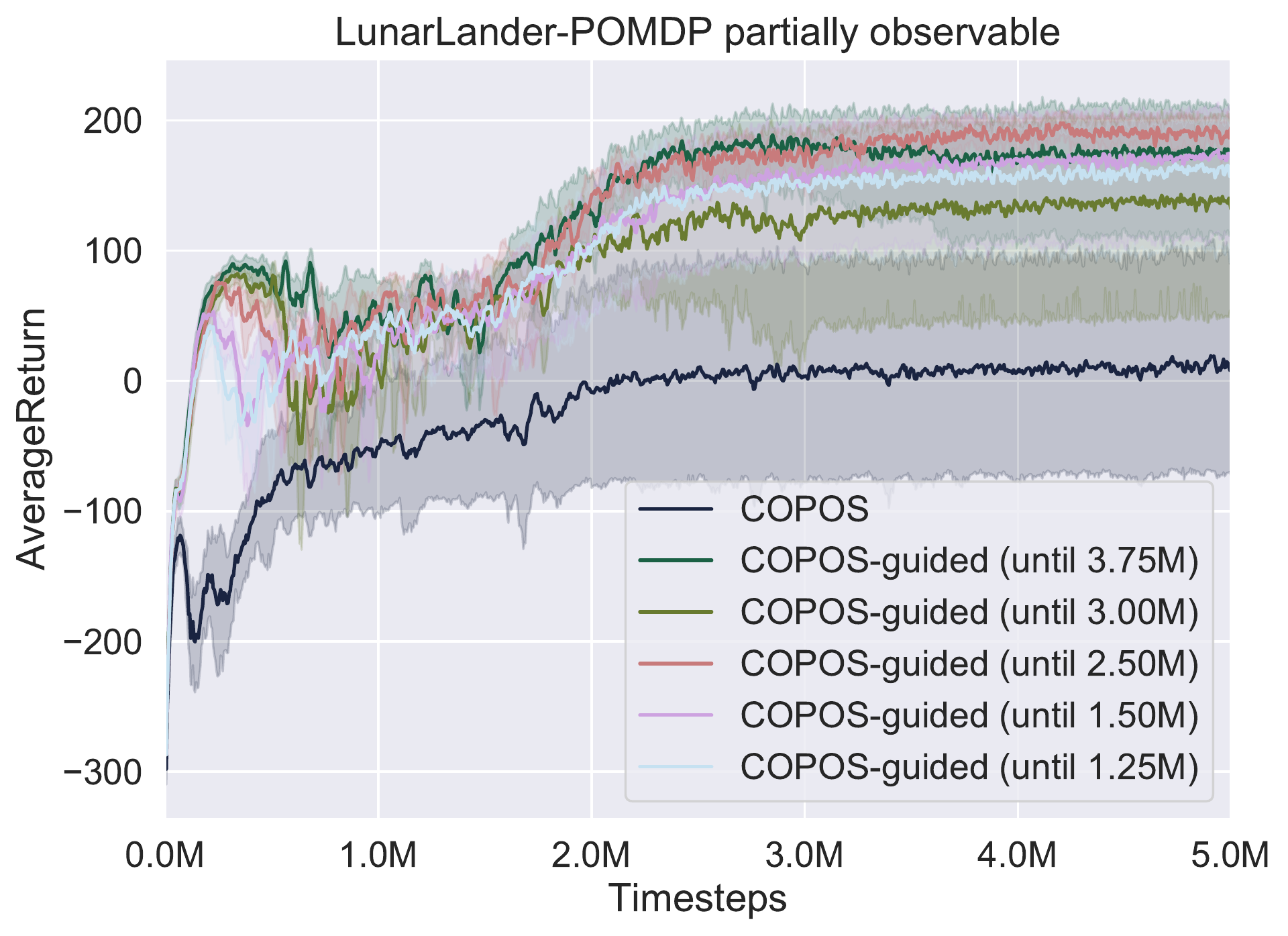}
  \caption{Average return for different mixing rates on
    LunarLander-POMDP with partial observations over 10 random
    seeds. In the legend, value in parenthesis indicates when mixing
    full and partial observations stops and purely partial
    observations are used.  Algorithms were executed for 5 million
    time steps. Shaded area denotes the bootstrapped 95\% confidence
    interval.}
  \label{fig:plot-lunarlander-mixing-param-sensitivity}
\end{figure}
Since the same mixing rate schedule of full and partial observations
worked in all the tasks our approach does not appear to be sensitive
to the mixing rate. To test the sensitivity to the mixing rate
schedules even further, we ran an additional experiment for
COPOS-guided with different schedules in LunarLander-POMDP as shown in
Figure~\ref{fig:plot-lunarlander-mixing-param-sensitivity}. For all
schedules COPOS-guided outperformed \gls{COPOS}.

\section{Real-Robot Experiment}
\begin{figure}[t]
\vspace{0.5em}
 \centering
 \setlength{\tabcolsep}{0.0pt}
\begin{tabular}{cc}
  \includegraphics[height=0.135\textheight]{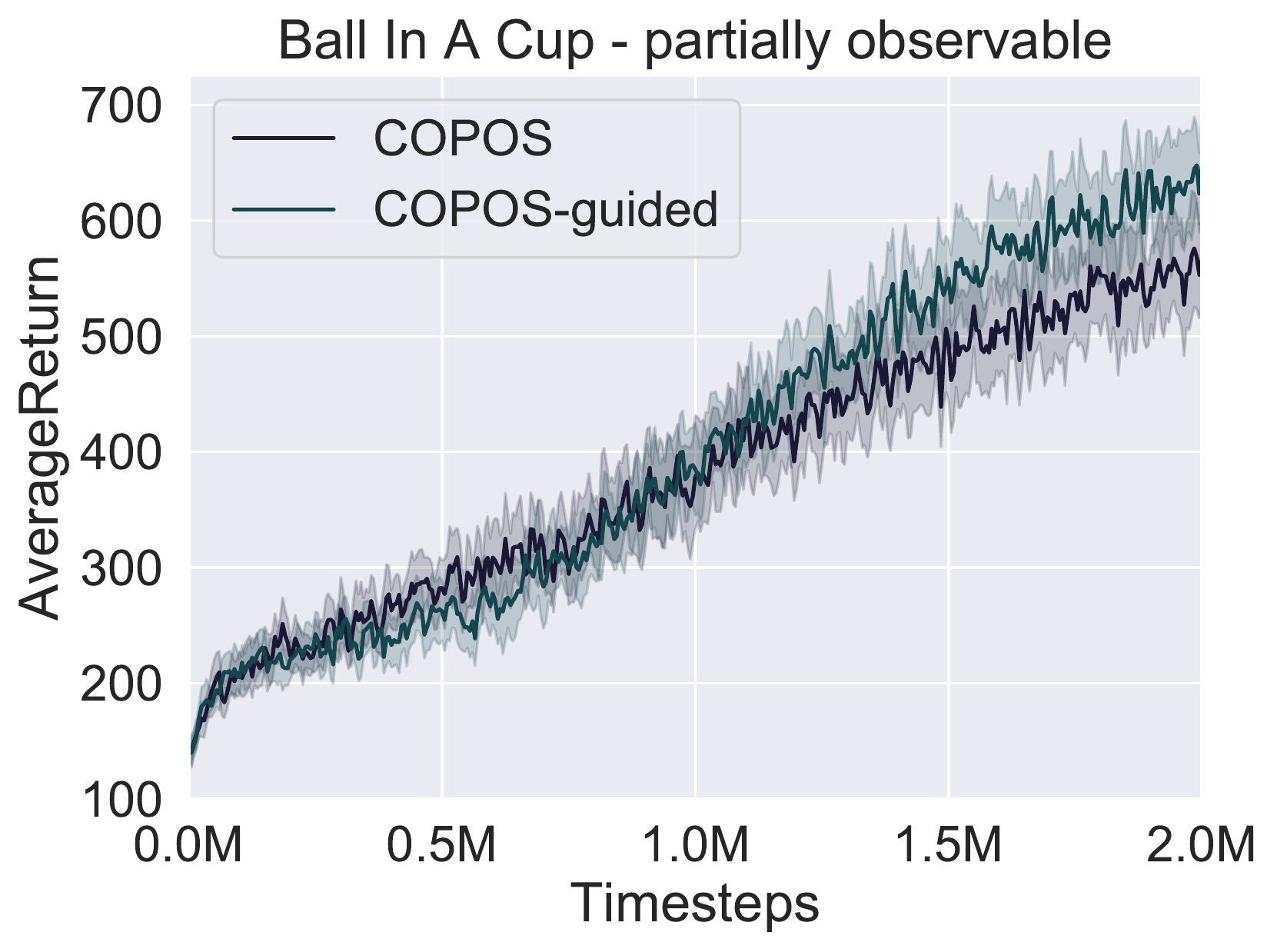} &
  \includegraphics[height=0.135\textheight]{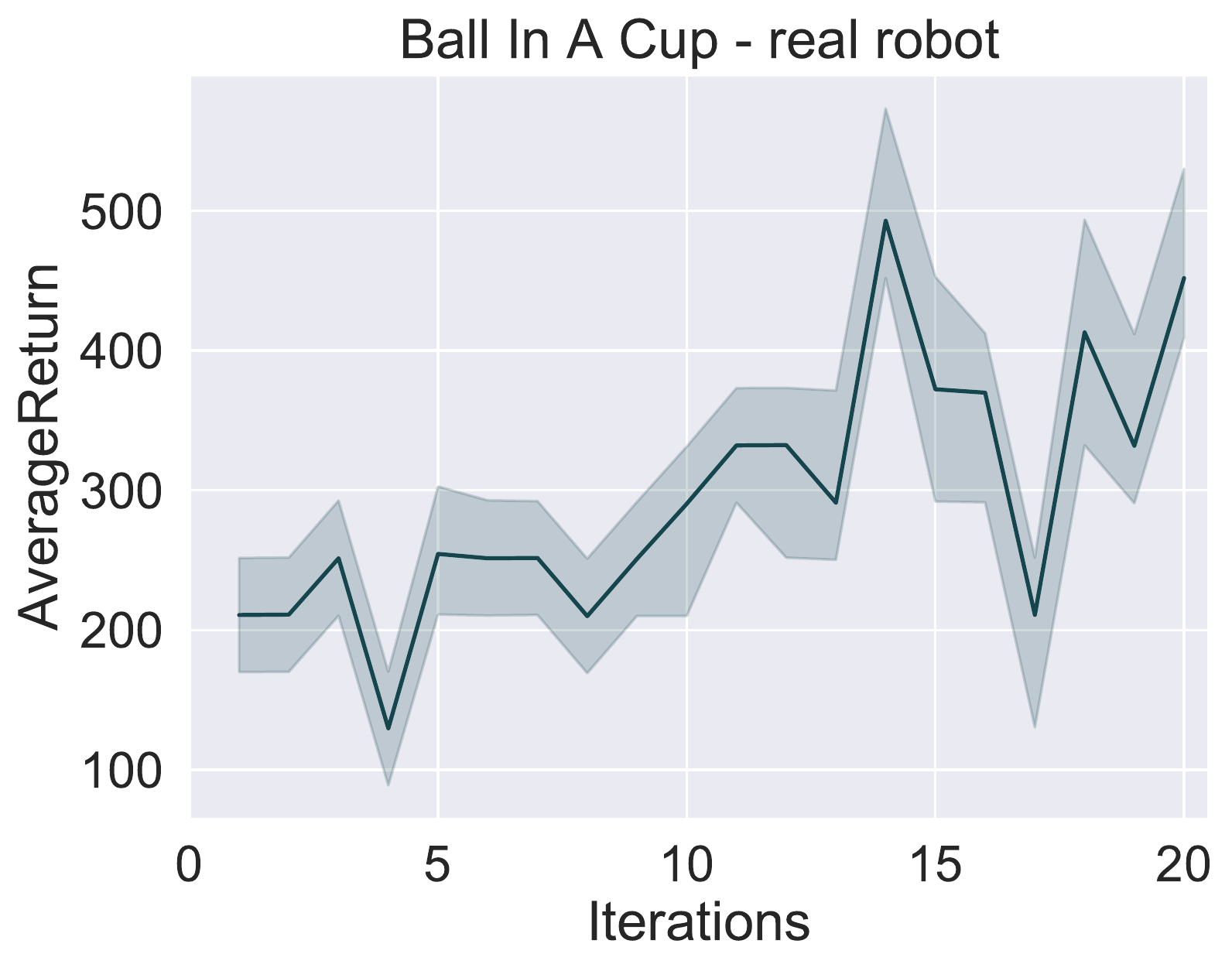}
  \end{tabular}
  \caption{Results for the ball-in-a-cup task using the Barrett WAM
    robot arm. The left plot shows the average return for COPOS and
    COPOS-guided in simulation runs. The right plot shows the average
    return obtained when continuing training of COPOS-guided on the
    real robot as well as the $0.25$ and $0.75$ computed via bootstrap sampling.}
\label{fig:real-robot}
\vspace{-15pt}
\end{figure}
To validate our method in a real-world sim2real scenario, we applied
\gls{GRL} using \gls{COPOS} to a ball in a cup task shown in
Fig.~\ref{fig:teaser} (top right). In the task a robot with a WAM arm
from Barrett Technologies tries to swing the ball hanging from a
string into a cup. In our setup, the actions of the RL agent are the
displacements of the desired joint positions that are then tracked by
the low-level controller of the robot. The agent controls only three
of the seven degrees of freedom of the robot, as the remaining DoFs
are not necessary to learn the task. As can be seen in
Fig.~\ref{fig:teaser} (top right), both ball and cup are equipped with
OptiTrack markers that allows to reliably estimate the position of the
ball and the cup. This allows to compute the reward function on the
real system and generate observations for the RL agent, which consist
of current joint position and velocity of the robot, desired joint
position as well as the position of the ball. Under partial
observability the velocity of the ball is not observed and at each
time step we freeze the pall position with $5$\% probability for 5
consecutive frames. We first train the policy using COPOS and
COPOS-guided in a simulator for $2$M time steps. Since COPOS-guided
outperformed COPOS in simulation as shown in Fig.~\ref{fig:real-robot}
(left) we continued with COPOS-guided for another $20$ iterations on
the real robot, where in each iteration $20$ episodes are
performed. As shown in Fig.~\ref{fig:real-robot} (right), this
learning step on the real system increased the average reward from
$210$ to $452$. Finally, we executed both the learned policy before
and after training on the real robot using $40$ episodes in order to
gain an impression of the qualitative performance of the policies. We
observed, that training on the real robot improved the success rate of
the policy from $17.5\%$ to $45\%$, underlining the potential of
exploiting state-information during training in simulation with a
subsequent training step in reality.

\section{Conclusion}
In this work, we proposed \gls{GRL} which guides \gls{RL} algorithms
with additional full state information during training to increase the
performance in the test phase. \gls{GRL} produces a policy which does
not rely on assumptions about state information in the test
phase. During training \gls{GRL} mixes samples with both full and
partial state information, starting with full observability but ending
with a policy needing only partial observations. The simple
formulation allows applying the approach to a variety of \gls{RL}
methods. Further, we are able to efficiently learn behavior for parts
of the problem that are actually fully observable while making
learning easier for the algorithms due to using full observability at
the beginning of learning. In the experiments, \gls{GRL} outperformed
baseline algorithms trained directly on partial observations in
different continuous and discrete tasks. As a history representation
needed in \gls{POMDP} tasks we used a simple windowing approach in
form of a fixed-length history. In future work it would be interesting
to replace this representation with an \gls{LSTM} to be able to
memorize events over a very long horizon. Our approach mixes fully and
partially observable samples linearly. At this point there is also
potential for future work by applying more complex functions for the
ratio between fully and partially observable samples at the beginning
of training.

\bibliographystyle{IEEEtran}
\balance
\bibliography{lit}

\begin{thebibliography}{10}
\providecommand{\url}[1]{#1}
\csname url@rmstyle\endcsname
\providecommand{\newblock}{\relax}
\providecommand{\bibinfo}[2]{#2}
\providecommand\BIBentrySTDinterwordspacing{\spaceskip=0pt\relax}
\providecommand\BIBentryALTinterwordstretchfactor{4}
\providecommand\BIBentryALTinterwordspacing{\spaceskip=\fontdimen2\font plus
\BIBentryALTinterwordstretchfactor\fontdimen3\font minus
  \fontdimen4\font\relax}
\providecommand\BIBforeignlanguage[2]{{%
\expandafter\ifx\csname l@#1\endcsname\relax
\typeout{** WARNING: IEEEtran.bst: No hyphenation pattern has been}%
\typeout{** loaded for the language `#1'. Using the pattern for}%
\typeout{** the default language instead.}%
\else
\language=\csname l@#1\endcsname
\fi
#2}}

\bibitem{silver2017mastering}
D.~Silver, J.~Schrittwieser, K.~Simonyan, I.~Antonoglou, A.~Huang, A.~Guez,
  T.~Hubert, L.~Baker, M.~Lai, A.~Bolton, \emph{et~al.}, ``Mastering the game
  of go without human knowledge,'' \emph{Nature}, vol. 550, no. 7676, p. 354,
  2017.

\bibitem{vinyals2019grandmaster}
O.~Vinyals, I.~Babuschkin, W.~M. Czarnecki, M.~Mathieu, A.~Dudzik, J.~Chung,
  D.~H. Choi, R.~Powell, T.~Ewalds, P.~Georgiev, \emph{et~al.}, ``Grandmaster
  level in starcraft ii using multi-agent reinforcement learning,''
  \emph{Nature}, vol. 575, no. 7782, pp. 350--354, 2019.

\bibitem{levine2016end}
S.~Levine, C.~Finn, T.~Darrell, and P.~Abbeel, ``End-to-end training of deep
  visuomotor policies,'' \emph{Journal of Machine Learning Research}, vol.~17,
  no.~1, pp. 1334--1373, 2016.

\bibitem{levine2018learning}
S.~Levine, P.~Pastor, A.~Krizhevsky, J.~Ibarz, and D.~Quillen, ``Learning
  hand-eye coordination for robotic grasping with deep learning and large-scale
  data collection,'' \emph{The International Journal of Robotics Research},
  vol.~37, no. 4-5, pp. 421--436, 2018.

\bibitem{kaelbling1998planning}
L.~P. Kaelbling, M.~L. Littman, and A.~R. Cassandra, ``Planning and acting in
  partially observable stochastic domains,'' \emph{Artificial intelligence},
  vol. 101, no. 1-2, pp. 99--134, 1998.

\bibitem{pajarinen2017robotic}
J.~Pajarinen and V.~Kyrki, ``Robotic manipulation of multiple objects as a
  {POMDP},'' \emph{Artificial Intelligence}, vol. 247, pp. 213--228, 2017.

\bibitem{radwan2018vlocnet++}
N.~Radwan, A.~Valada, and W.~Burgard, ``Vlocnet++: Deep multitask learning for
  semantic visual localization and odometry,'' \emph{IEEE Robotics and
  Automation Letters}, vol.~3, no.~4, pp. 4407--4414, 2018.

\bibitem{bakker2002reinforcement}
B.~Bakker, ``Reinforcement learning with long short-term memory,'' in
  \emph{Advances in neural information processing systems}, 2002, pp.
  1475--1482.

\bibitem{wierstra2007solving}
D.~Wierstra, A.~Foerster, J.~Peters, and J.~Schmidhuber, ``Solving deep memory
  pomdps with recurrent policy gradients,'' in \emph{International Conference
  on Artificial Neural Networks}.\hskip 1em plus 0.5em minus 0.4em\relax
  Springer, 2007, pp. 697--706.

\bibitem{hausknecht2015deep}
M.~Hausknecht and P.~Stone, ``Deep recurrent q-learning for partially
  observable mdps,'' in \emph{2015 AAAI Fall Symposium Series}, 2015.

\bibitem{heess2015memory}
N.~Heess, J.~J. Hunt, T.~P. Lillicrap, and D.~Silver, ``Memory-based control
  with recurrent neural networks,'' \emph{arXiv preprint arXiv:1512.04455},
  2015.

\bibitem{zhang2016learningmpc}
T.~Zhang, G.~Kahn, S.~Levine, and P.~Abbeel, ``Learning deep control policies
  for autonomous aerial vehicles with mpc-guided policy search,'' in \emph{IEEE
  International Conference on Robotics and Automation (ICRA)}.\hskip 1em plus
  0.5em minus 0.4em\relax IEEE, 2016, pp. 528--535.

\bibitem{levine2013guided}
S.~Levine and V.~Koltun, ``Guided policy search,'' in \emph{International
  Conference on Machine Learning}, 2013, pp. 1--9.

\bibitem{levine2014learningcomplex}
------, ``Learning complex neural network policies with trajectory
  optimization,'' in \emph{International Conference on Machine Learning}, 2014,
  pp. 829--837.

\bibitem{huttenrauch2017guided}
M.~H{\"u}ttenrauch, A.~{\v{S}}o{\v{s}}i{\'c}, and G.~Neumann, ``Guided deep
  reinforcement learning for swarm systems,'' 2017.

\bibitem{pinto2018}
L.~Pinto, M.~Andrychowicz, P.~Welinder, W.~Zaremba, and P.~Abbeel, ``Asymmetric
  actor critic for image-based robot learning,'' in \emph{Proceedings of
  Robotics: Science and Systems}, Pittsburgh, Pennsylvania, June 2018.

\bibitem{kaelbling1996reinforcement}
L.~P. Kaelbling, M.~L. Littman, and A.~W. Moore, ``Reinforcement learning: A
  survey,'' \emph{Journal of artificial intelligence research}, vol.~4, pp.
  237--285, 1996.

\bibitem{wierstra2007policy}
D.~Wierstra and J.~Schmidhuber, ``Policy gradient critics,'' in \emph{European
  Conference on Machine Learning}.\hskip 1em plus 0.5em minus 0.4em\relax
  Springer, 2007, pp. 466--477.

\bibitem{futoma2018prediction}
J.~Futoma, S.~Harvard, M.~C. Hughes, and F.~Doshi-Velez,
  ``Prediction-constrained pomdps,'' \emph{32nd Conference on Neural
  Information Processing Systems (NIPS)}, 2018.

\bibitem{igl2018deep}
M.~Igl, L.~Zintgraf, T.~A. Le, F.~Wood, and S.~Whiteson, ``Deep variational
  reinforcement learning for {POMDP}s,'' in \emph{International Conference on
  Machine Learning}, 2018, pp. 2122--2131.

\bibitem{levine1501learning}
S.~Levine, N.~Wagener, and P.~Abbeel, ``Learning contact-rich manipulation
  skills with guided policy search (2015),'' \emph{arXiv preprint
  arXiv:1501.05611}, 2015.

\bibitem{montgomery2016guided}
W.~H. Montgomery and S.~Levine, ``Guided policy search via approximate mirror
  descent,'' in \emph{Advances in Neural Information Processing Systems}, 2016,
  pp. 4008--4016.

\bibitem{zhang2016learning}
M.~Zhang, Z.~McCarthy, C.~Finn, S.~Levine, and P.~Abbeel, ``Learning deep
  neural network policies with continuous memory states,'' in \emph{2016 IEEE
  International Conference on Robotics and Automation (ICRA)}.\hskip 1em plus
  0.5em minus 0.4em\relax IEEE, 2016, pp. 520--527.

\bibitem{shani2013survey}
G.~Shani, J.~Pineau, and R.~Kaplow, ``A survey of point-based {POMDP}
  solvers,'' \emph{Autonomous Agents and Multi-Agent Systems}, vol.~27, no.~1,
  pp. 1--51, 2013.

\bibitem{wiering2012reinforcement}
M.~Wiering and M.~Van~Otterlo, ``Reinforcement learning,'' \emph{Adaptation,
  learning, and optimization}, vol.~12, p.~3, 2012.

\bibitem{schulman2015trust}
J.~Schulman, S.~Levine, P.~Abbeel, M.~Jordan, and P.~Moritz, ``Trust region
  policy optimization,'' in \emph{International conference on machine
  learning}, 2015, pp. 1889--1897.

\bibitem{schulman2017proximal}
J.~Schulman, F.~Wolski, P.~Dhariwal, A.~Radford, and O.~Klimov, ``Proximal
  policy optimization algorithms,'' \emph{arXiv preprint arXiv:1707.06347},
  2017.

\bibitem{baselines}
P.~Dhariwal, C.~Hesse, O.~Klimov, A.~Nichol, M.~Plappert, A.~Radford,
  J.~Schulman, S.~Sidor, Y.~Wu, and P.~Zhokhov, ``Openai baselines,''
  \url{https://github.com/openai/baselines}, 2017.

\bibitem{stable-baselines}
A.~Hill, A.~Raffin, M.~Ernestus, A.~Gleave, R.~Traore, P.~Dhariwal, C.~Hesse,
  O.~Klimov, A.~Nichol, M.~Plappert, A.~Radford, J.~Schulman, S.~Sidor, and
  Y.~Wu, ``Stable baselines,''
  \url{https://github.com/hill-a/stable-baselines}, 2018.

\bibitem{rongzhi}
R.~Zhi, ``Deep reinforcement learning under uncertainty for autonomous
  driving,'' Master's thesis, Technische Universit{\"a}t Darmstadt, 2018.

\bibitem{todorov2012mujoco}
E.~Todorov, T.~Erez, and Y.~Tassa, ``Mujoco: A physics engine for model-based
  control,'' in \emph{2012 IEEE/RSJ International Conference on Intelligent
  Robots and Systems}.\hskip 1em plus 0.5em minus 0.4em\relax IEEE, 2012, pp.
  5026--5033.

\bibitem{openaigym}
G.~Brockman, V.~Cheung, L.~Pettersson, J.~Schneider, J.~Schulman, J.~Tang, and
  W.~Zaremba, ``Openai gym,'' 2016.

\bibitem{smith2004heuristic}
T.~Smith and R.~Simmons, ``Heuristic search value iteration for pomdps,'' in
  \emph{Proceedings of the 20th conference on Uncertainty in artificial
  intelligence}.\hskip 1em plus 0.5em minus 0.4em\relax AUAI Press, 2004, pp.
  520--527.

\bibitem{gympomdp}
S.~Totaro, ``gym-pomdp,'' \url{https://github.com/d3sm0/gym_pomdp}, 2018.

\bibitem{hochreiter1997long}
S.~Hochreiter and J.~Schmidhuber, ``Long short-term memory,'' \emph{Neural
  computation}, vol.~9, no.~8, pp. 1735--1780, 1997.

\end{thebibliography}

\onecolumn
\section*{Supplementary Material}

\section{Additional Plots for Experiments}
\begin{figure}[ht]
			\centering
			\begin{subfigure}[b]{0.47\columnwidth}
				\centering
				\includegraphics[width=\columnwidth]{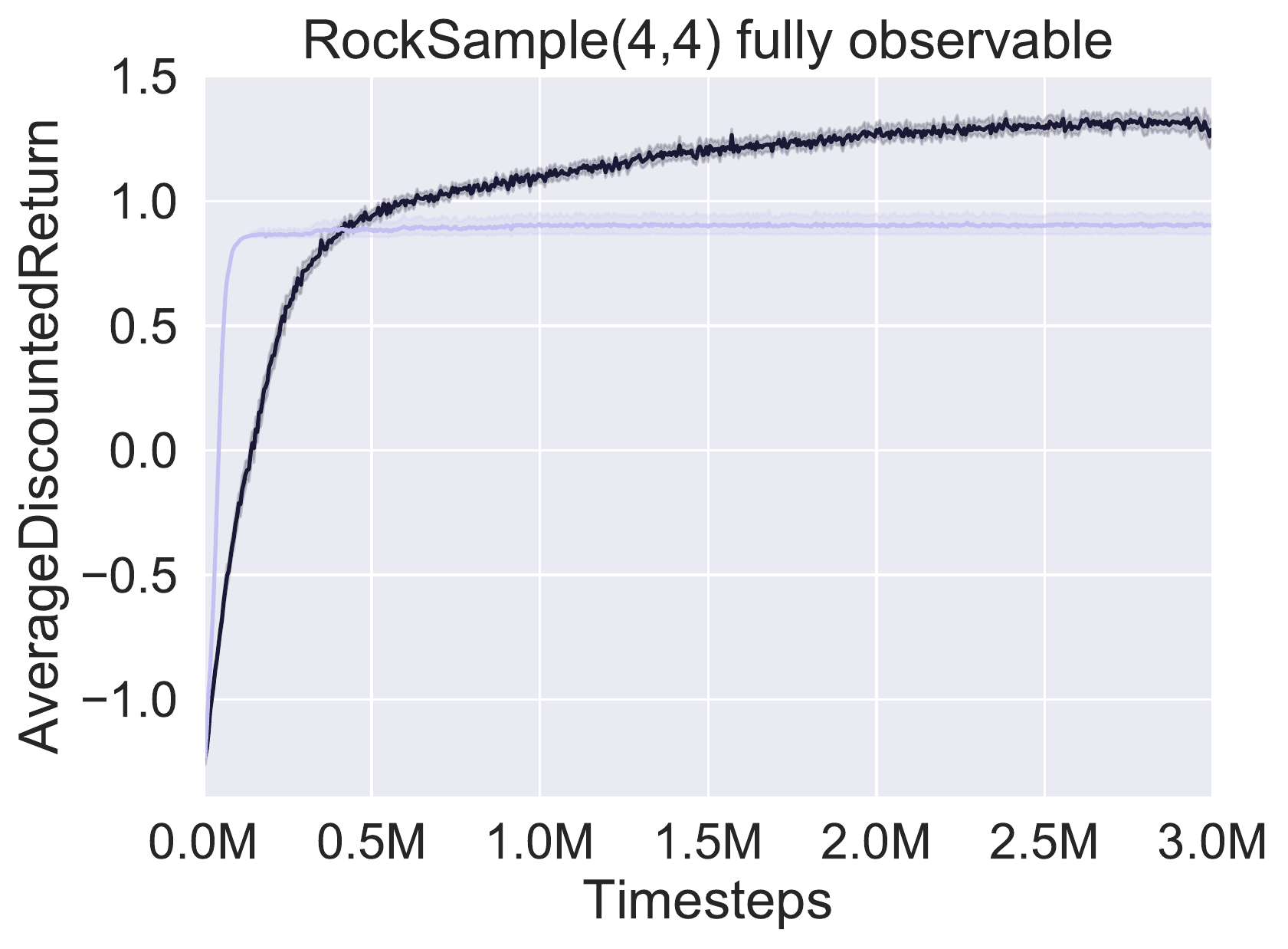}
				\label{fig:plot-rocksample4x4-full-return}
			\end{subfigure}
			\hfill
			\begin{subfigure}[b]{0.47\columnwidth}
				\centering
				\includegraphics[width=\columnwidth]{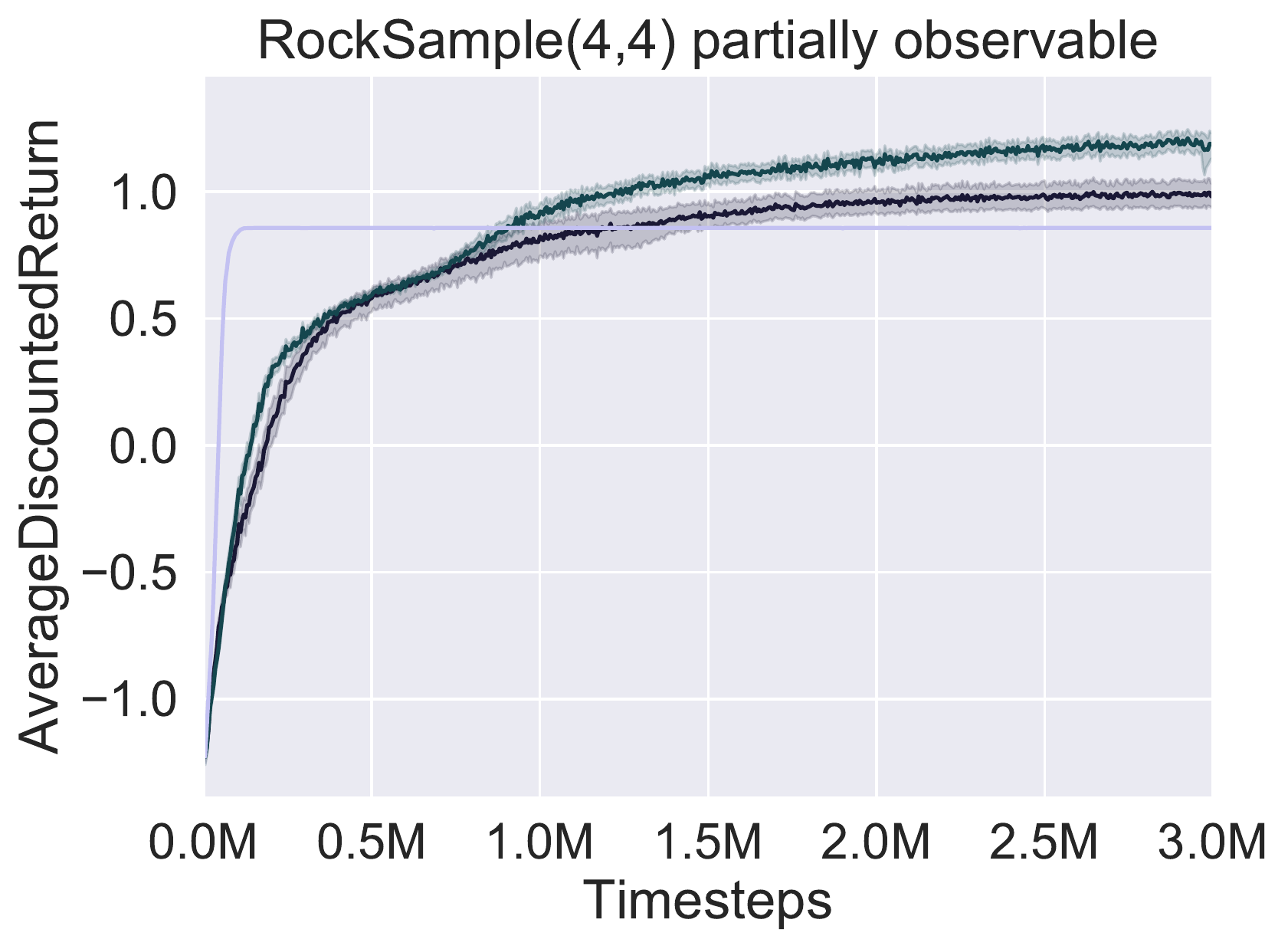}
				\label{fig:plot-rocksample4x4-pomdp-return}
			\end{subfigure}
			\bigskip
			\begin{subfigure}[b]{0.47\columnwidth}
				\centering
				\includegraphics[width=\columnwidth]{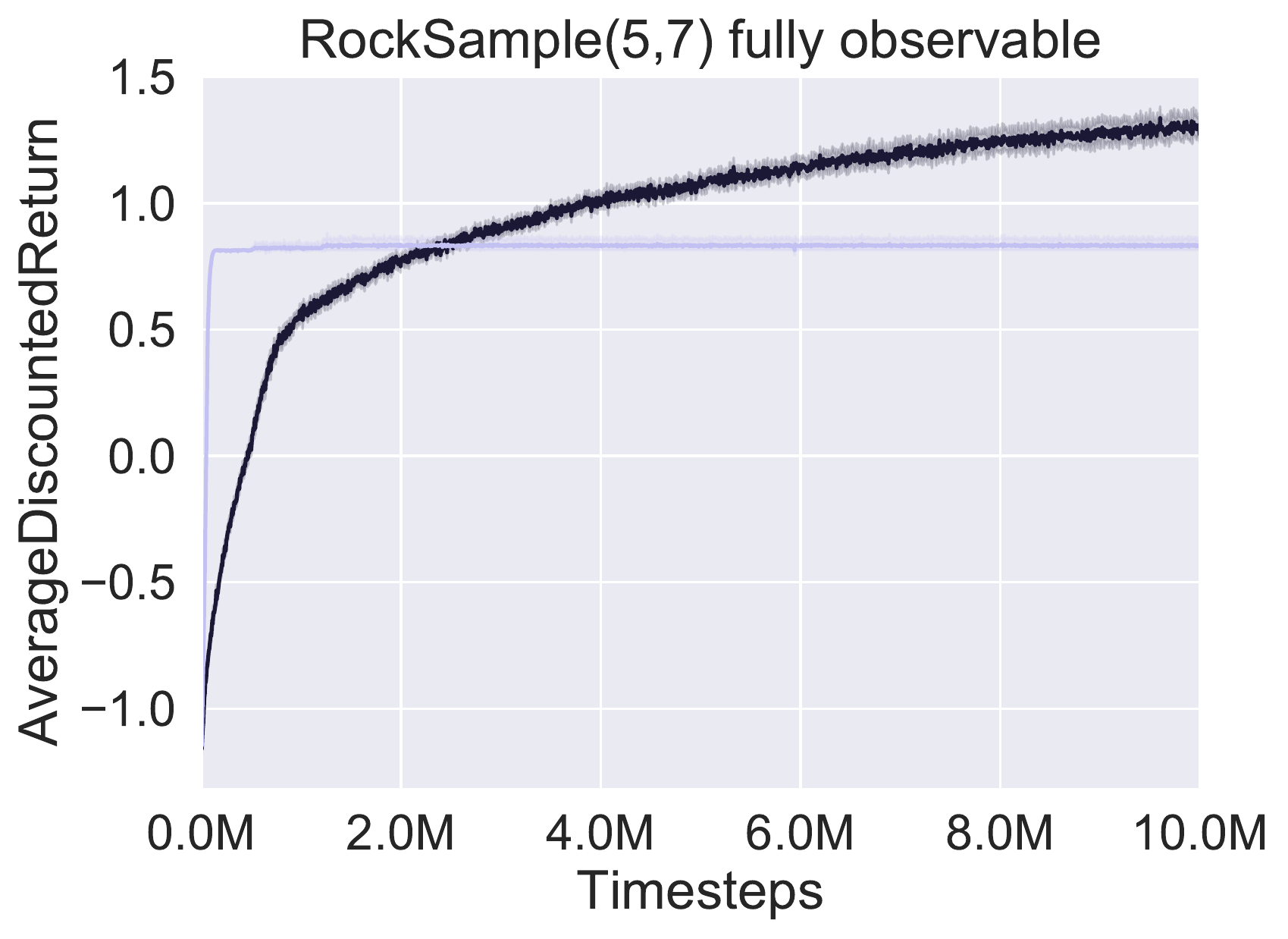}
				\label{fig:plot-rocksample5x7-full-return}
			\end{subfigure}
			\hfill
			\begin{subfigure}[b]{0.47\columnwidth}
				\centering
				\includegraphics[width=\columnwidth]{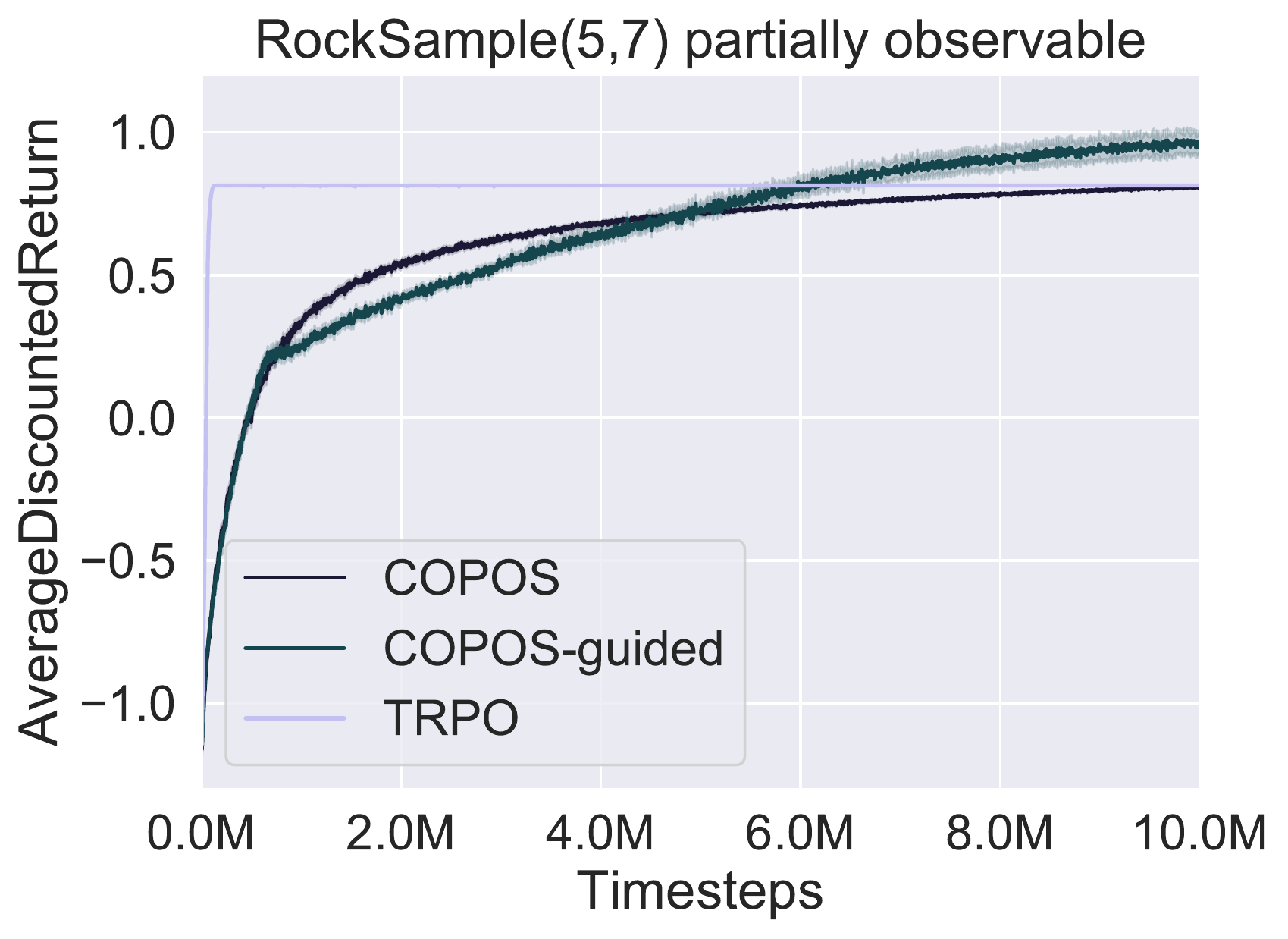}
				\label{fig:plot-rocksample5x7-pomdp-return}
			\end{subfigure}
			\caption{Average discounted return for two instances of RockSample, both with full observations (left) and partial observations (right) over 50 random seeds. Algorithms were executed for 5 million time steps on RockSample(4,4) (top) and 10 million time steps on RockSample(5,7) (bottom). Shaded area denotes the bootstrapped 95\% confidence interval.}
			\label{fig:plot-rocksample-return}
\end{figure}

\begin{figure}[ht]
	\centering
			\begin{subfigure}[b]{0.47\columnwidth}
				\centering
				\includegraphics[width=\columnwidth]{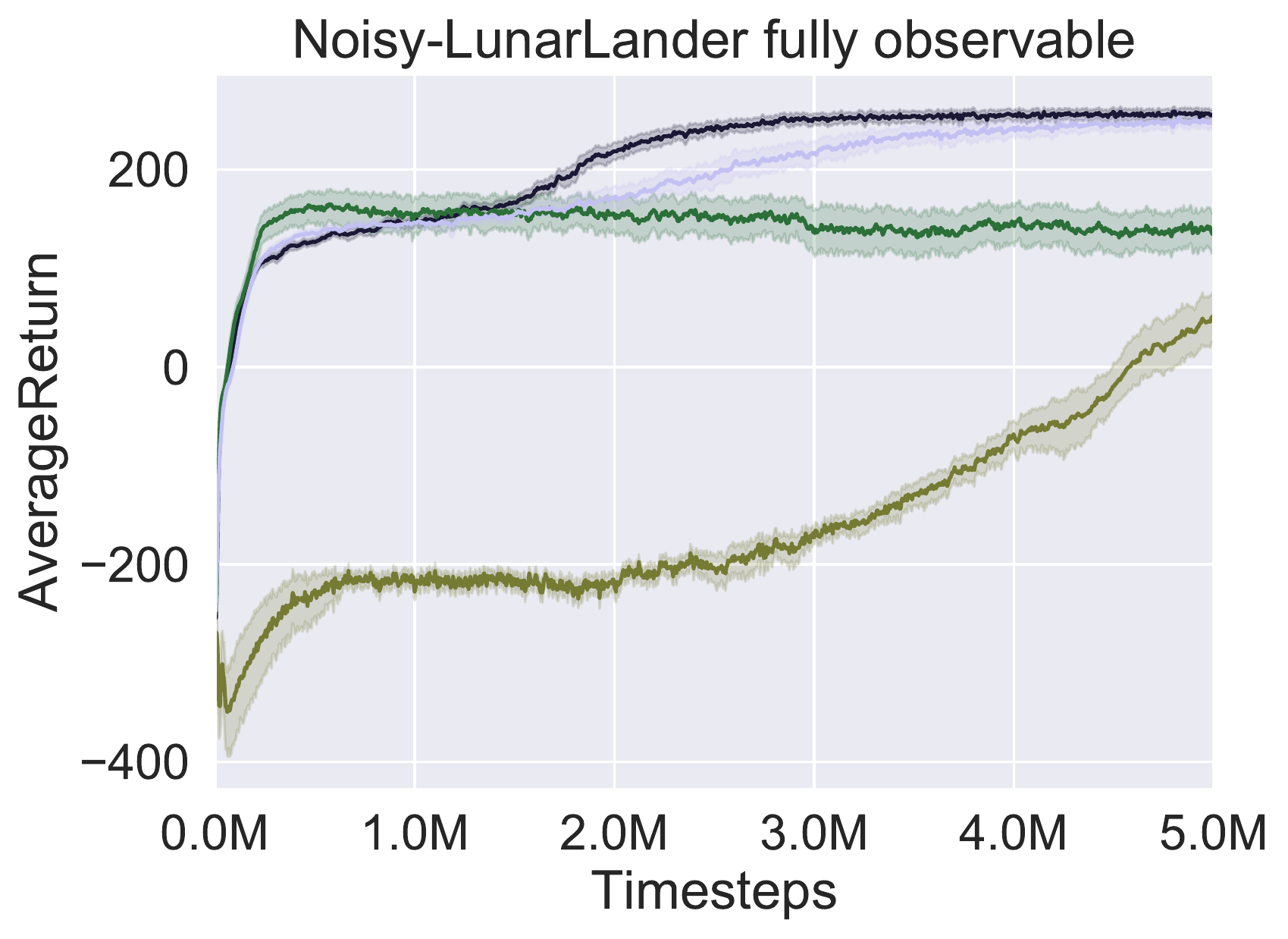}
			\end{subfigure}
			\hfill
			\begin{subfigure}[b]{0.47\columnwidth}
				\centering
				\includegraphics[width=\columnwidth]{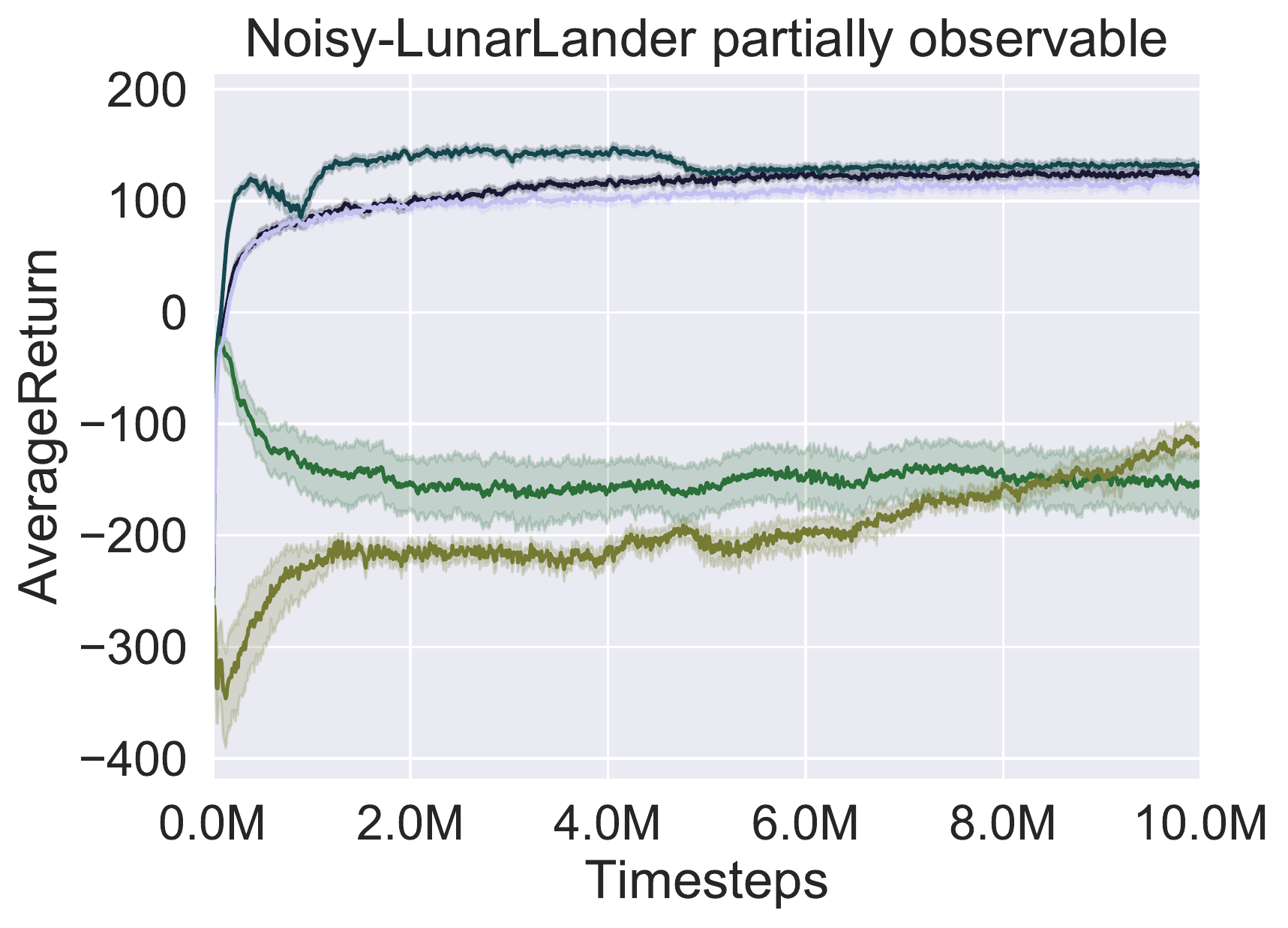}
			\end{subfigure}
			\bigskip
			\begin{subfigure}[b]{0.47\columnwidth}
				\centering
				\includegraphics[width=\columnwidth]{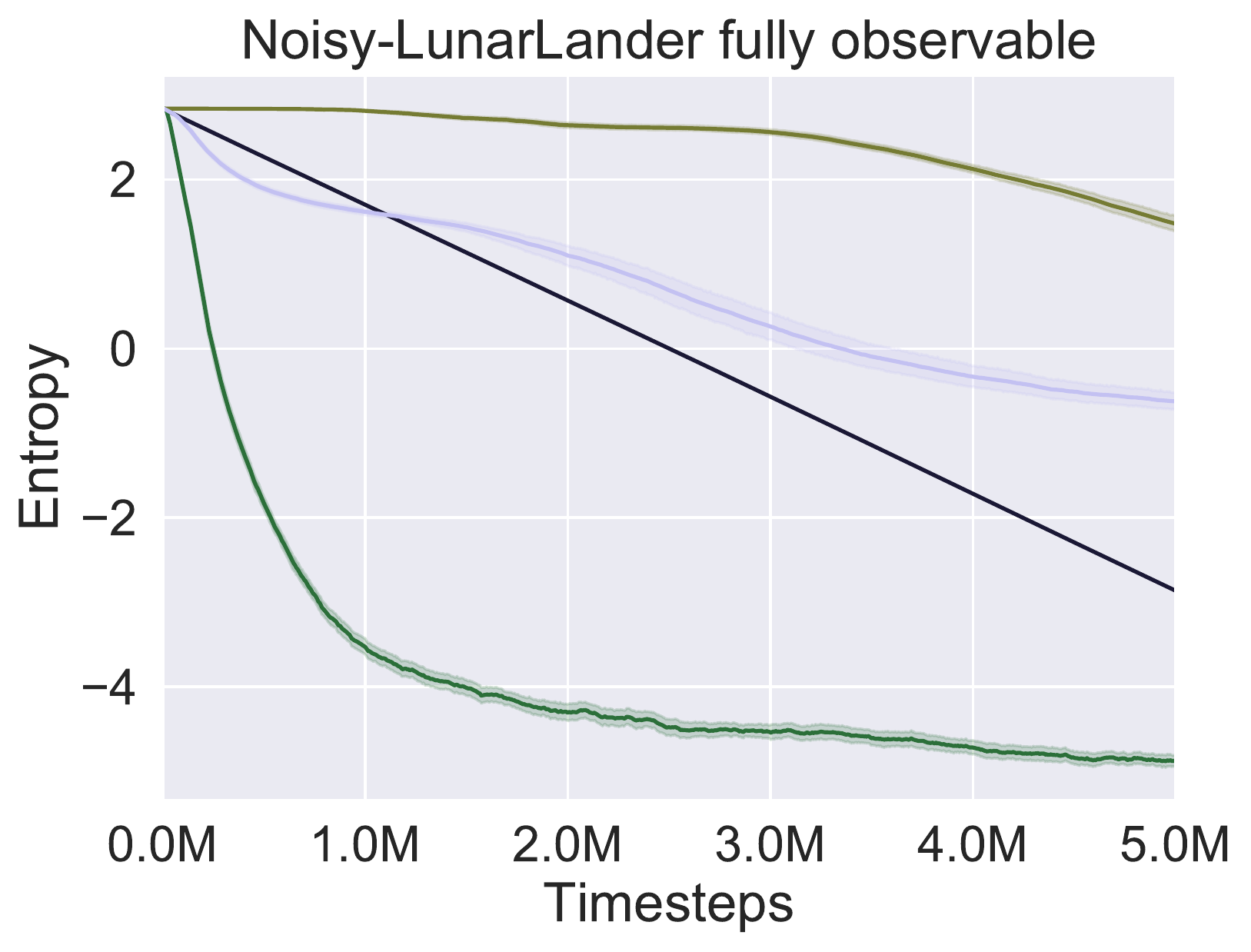}
			\end{subfigure}
			\hfill
			\begin{subfigure}[b]{0.47\columnwidth}
				\centering
				\includegraphics[width=\columnwidth]{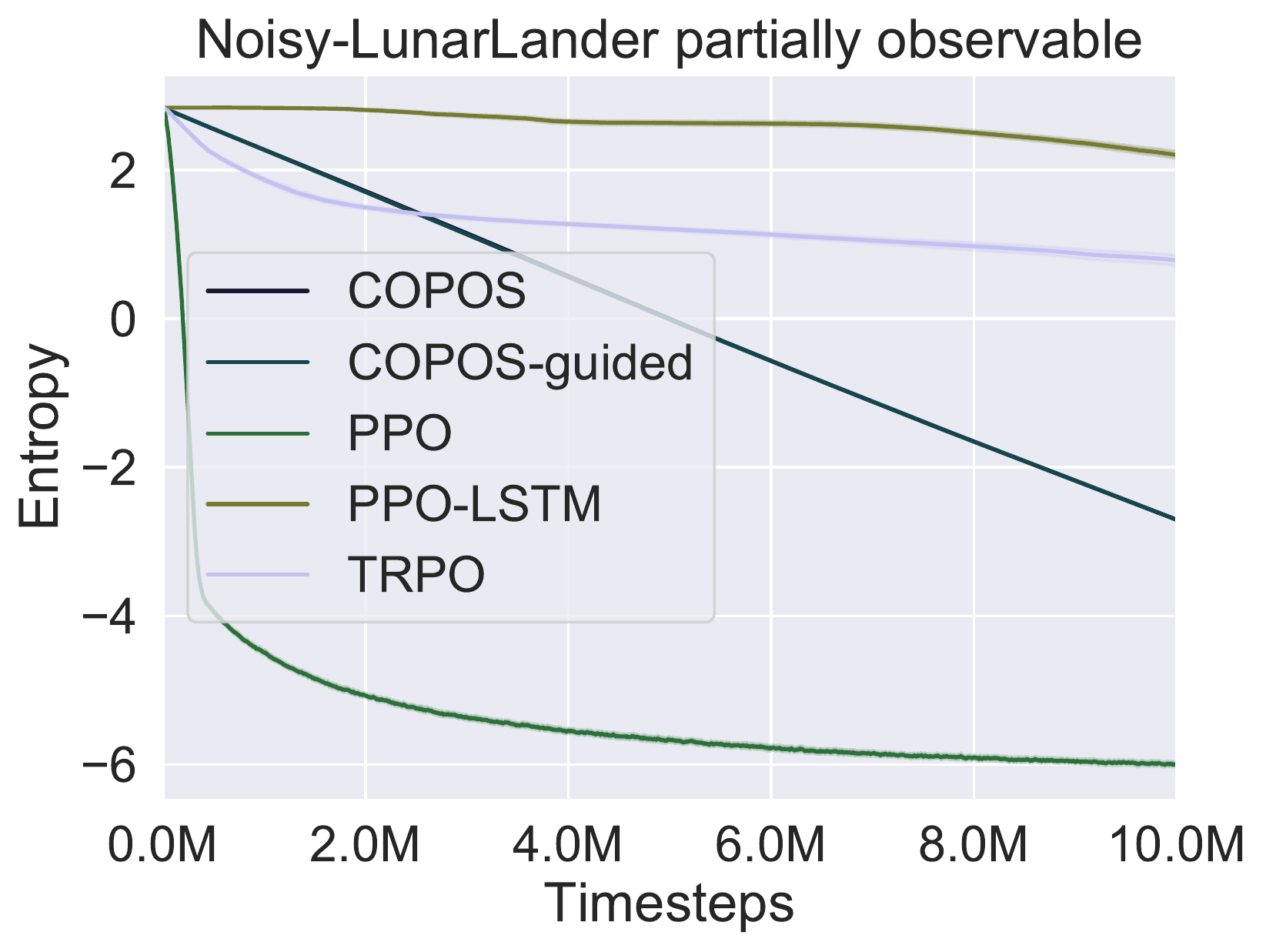}
			\end{subfigure}
			\caption{Average return (top) and entropy (bottom) for both Noisy-LunarLander with full observations (left) and noisy observations (right) over 50 random seeds. Algorithms were executed for 10 million time steps (5 million time steps in the fully observable case). Shaded area denotes the bootstrapped 95\% confidence interval.}
			\label{fig:plot-noisylunarlander}
\end{figure}

\begin{figure*}[ht!]
	\centering
			\begin{subfigure}[b]{0.32\textwidth}
				\centering
				\includegraphics[width=\textwidth]{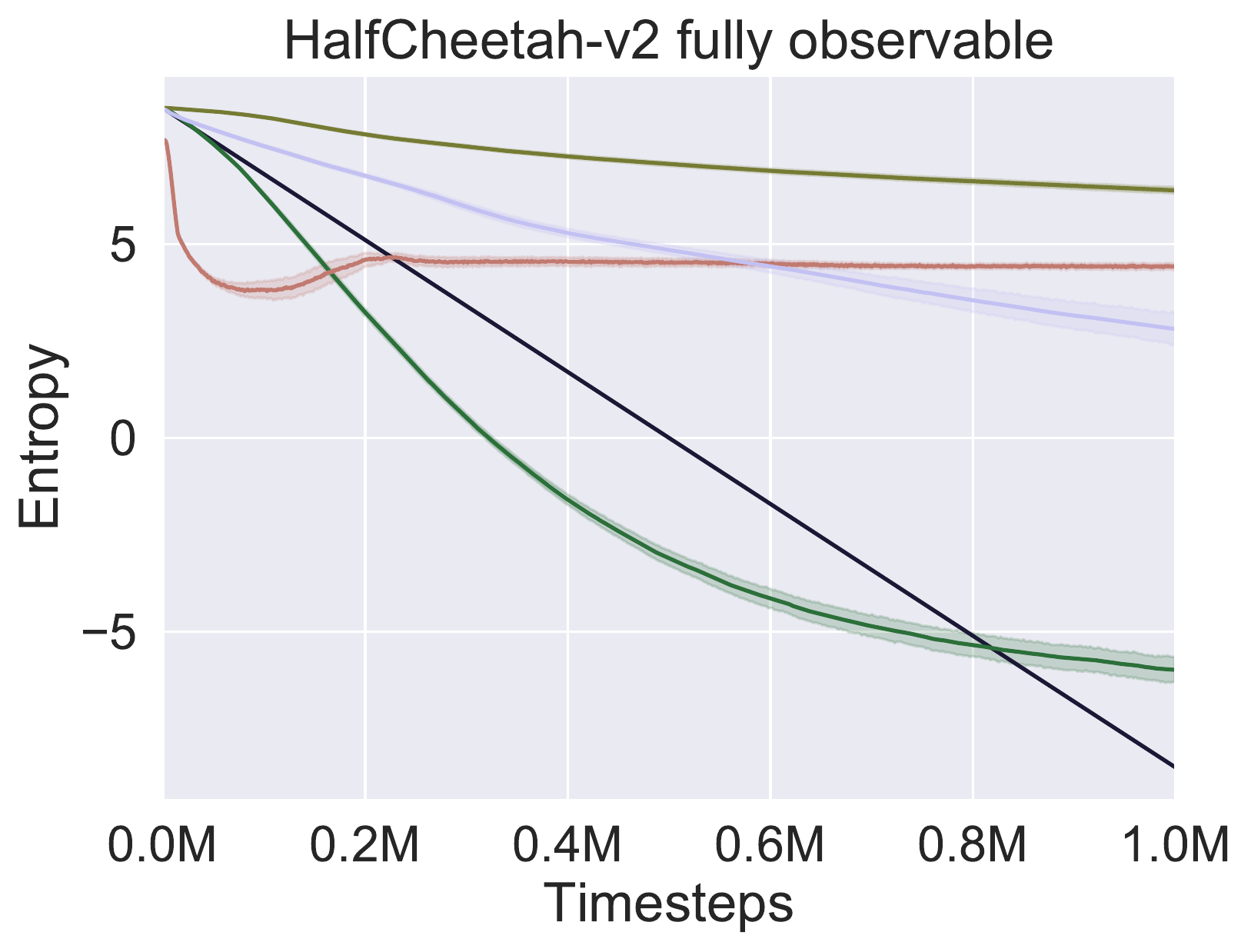}
			\end{subfigure}
			\hfill
			\begin{subfigure}[b]{0.32\textwidth}
				\centering
				\includegraphics[width=\textwidth]{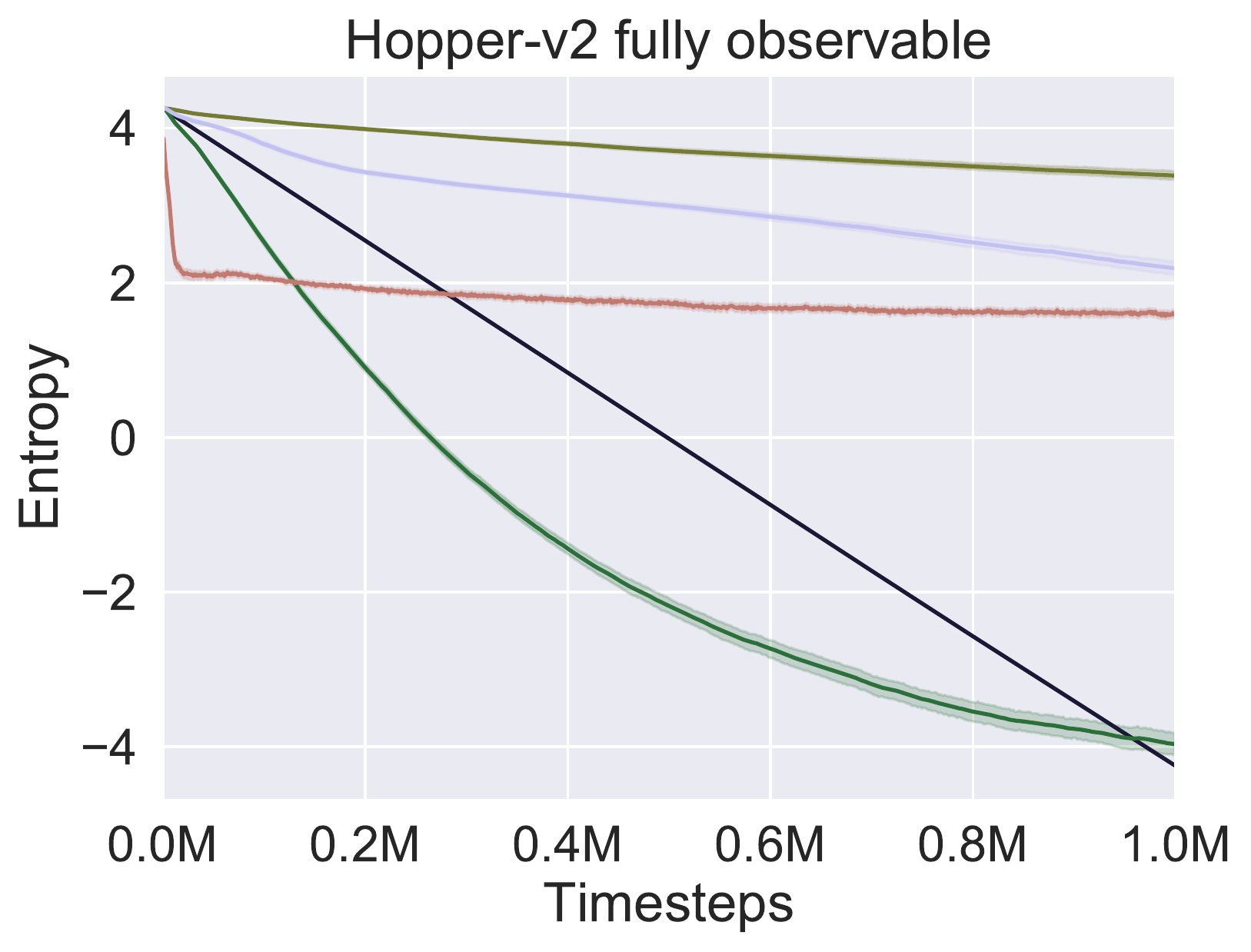}
			\end{subfigure}
			\hfill
			\begin{subfigure}[b]{0.32\textwidth}
				\centering
				\includegraphics[width=\textwidth]{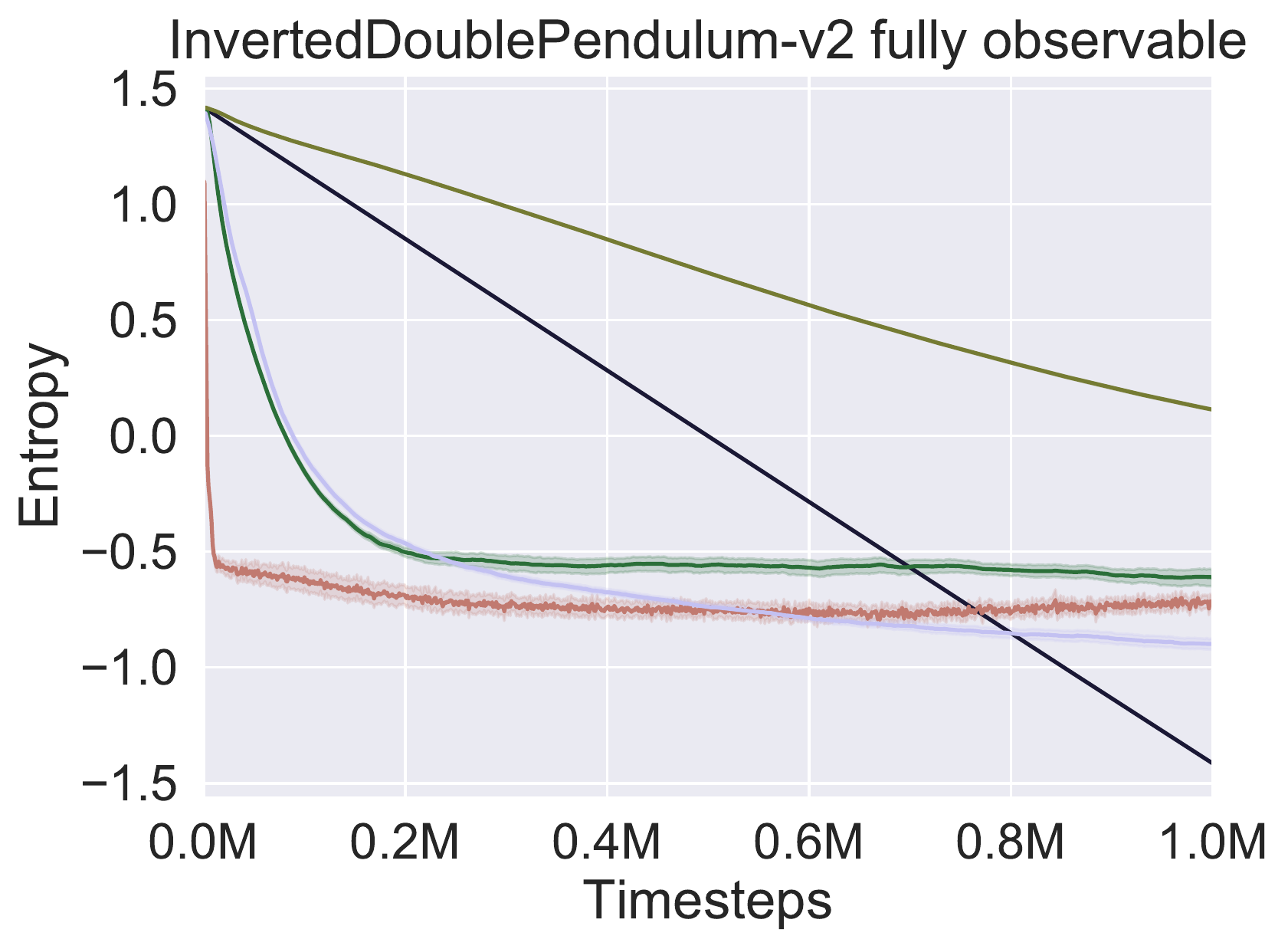}
			\end{subfigure}
			\bigskip
			\begin{subfigure}[b]{0.32\textwidth}
				\centering
				\includegraphics[width=\textwidth]{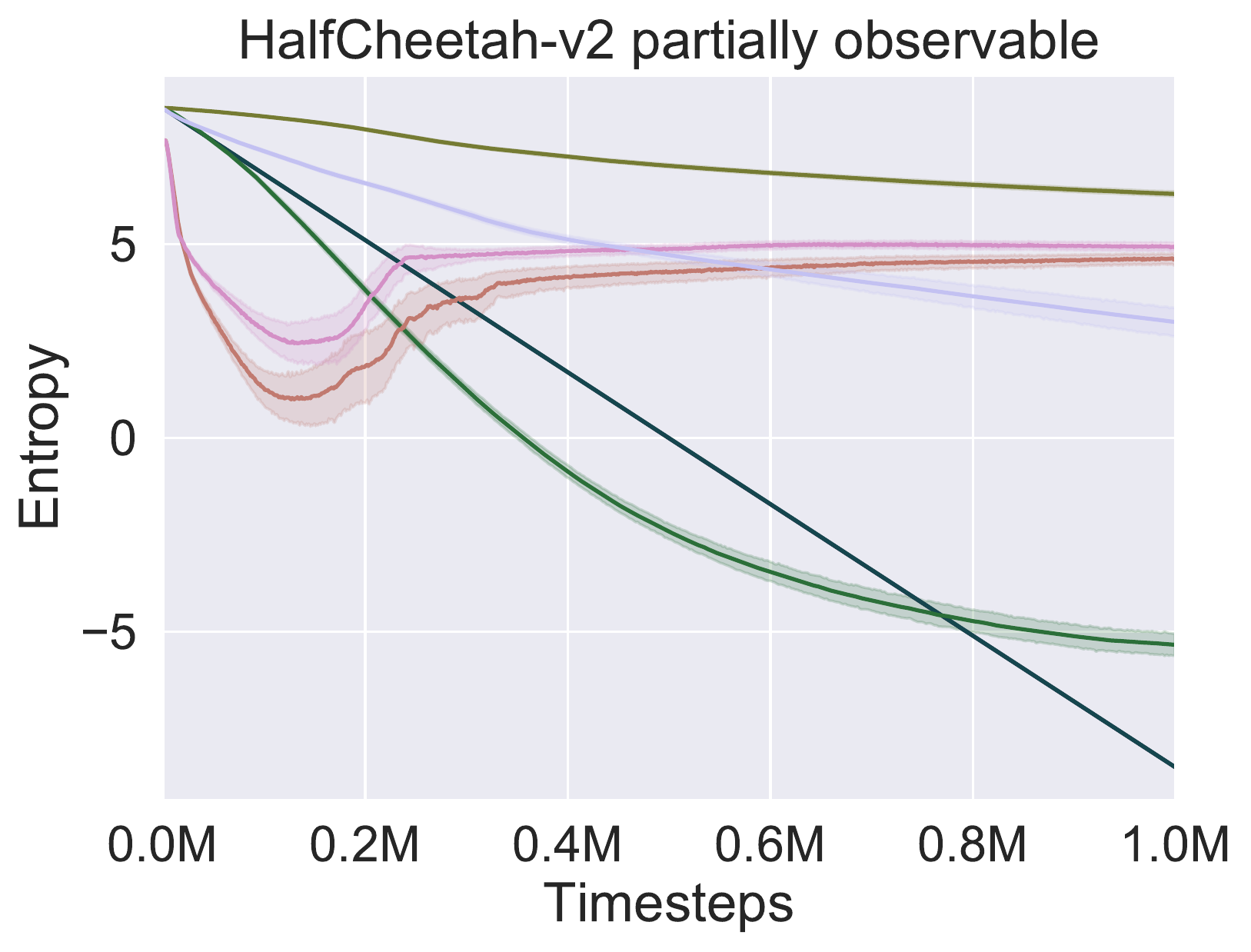}
			\end{subfigure}
			\hfill
			\begin{subfigure}[b]{0.32\textwidth}
				\centering
				\includegraphics[width=\textwidth]{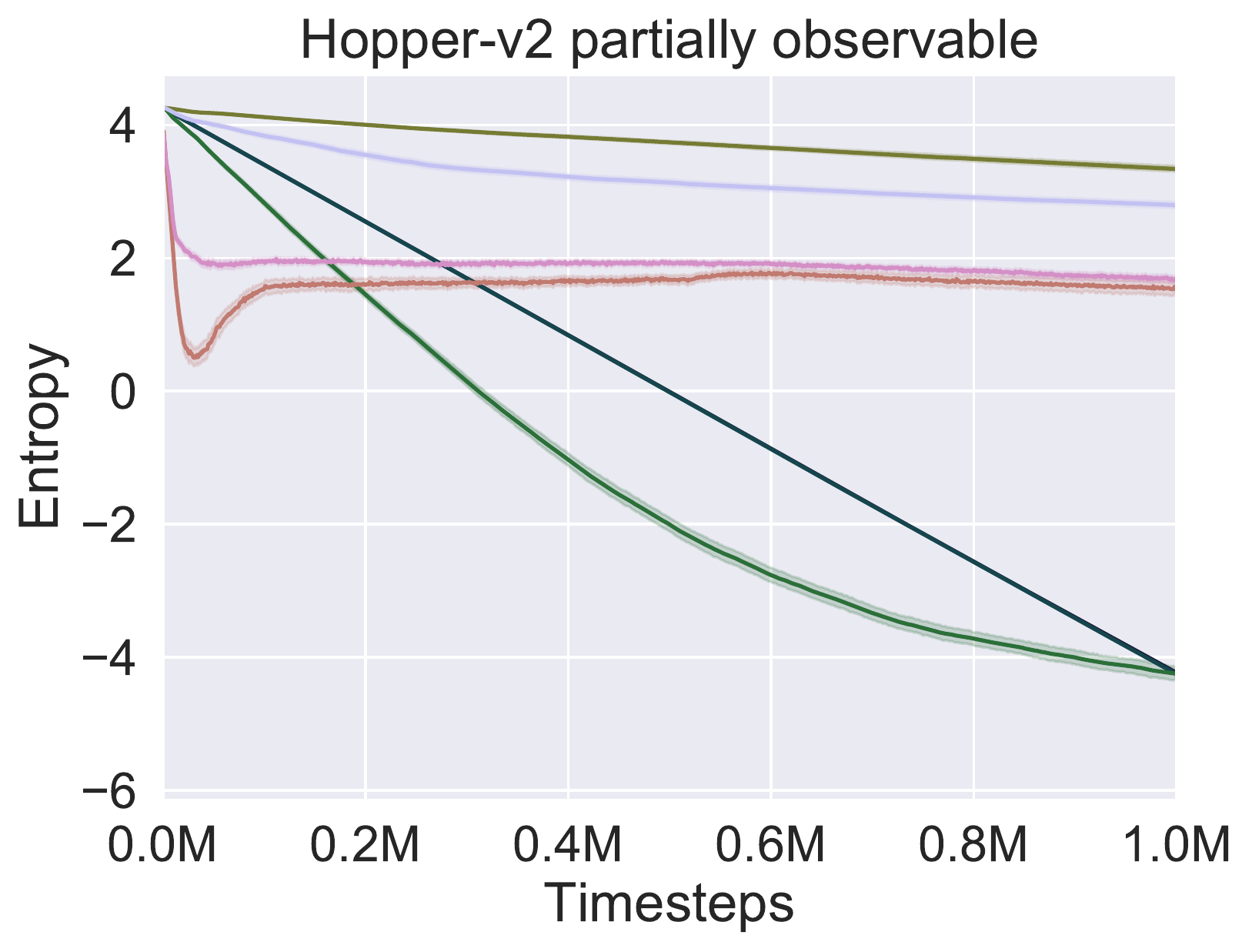}
			\end{subfigure}
			\hfill
			\begin{subfigure}[b]{0.32\textwidth}
				\centering
				\includegraphics[width=\textwidth]{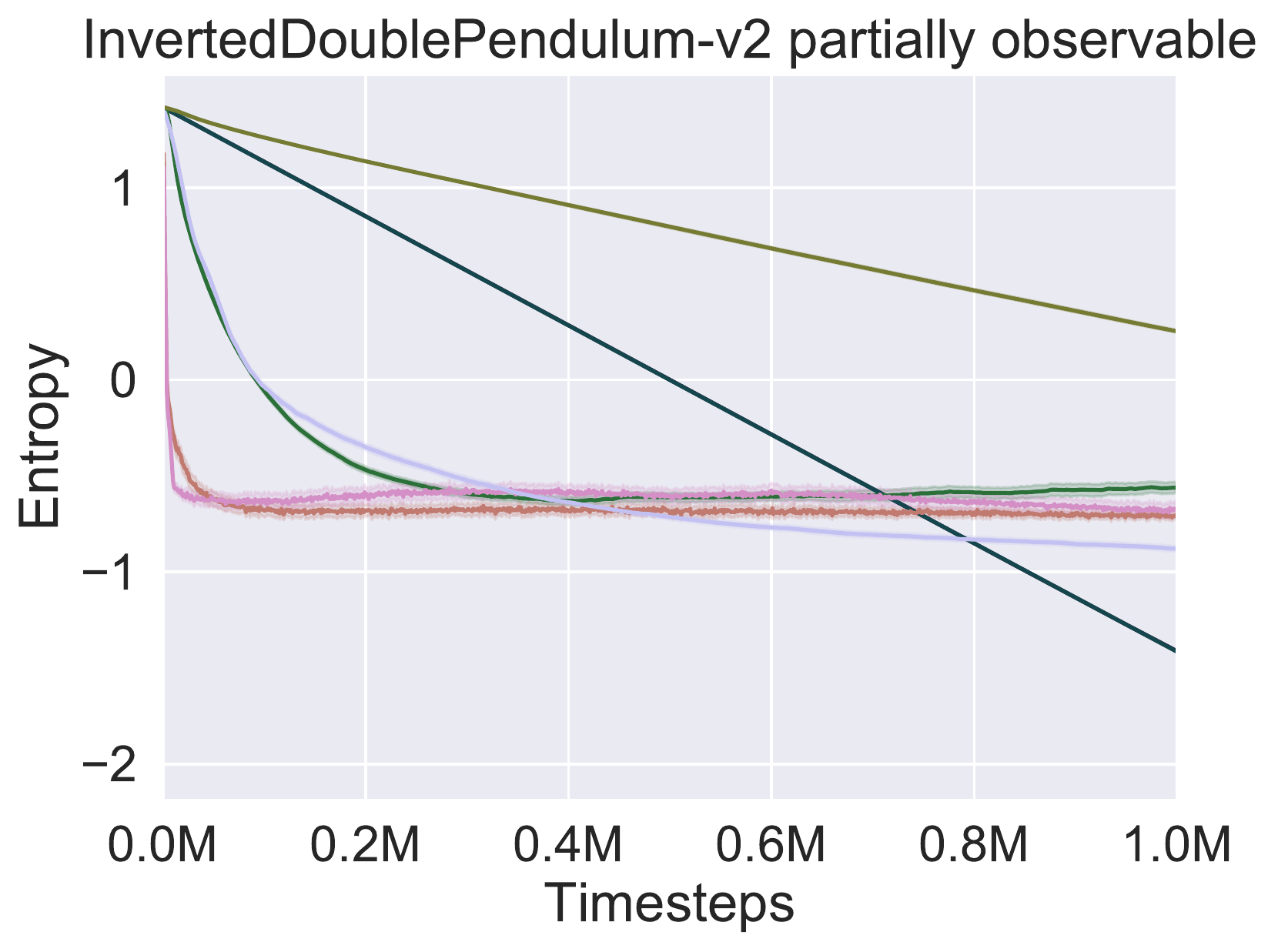}
			\end{subfigure}
			\bigskip
			\begin{subfigure}[b]{0.32\textwidth}
				\centering
				\includegraphics[width=\textwidth]{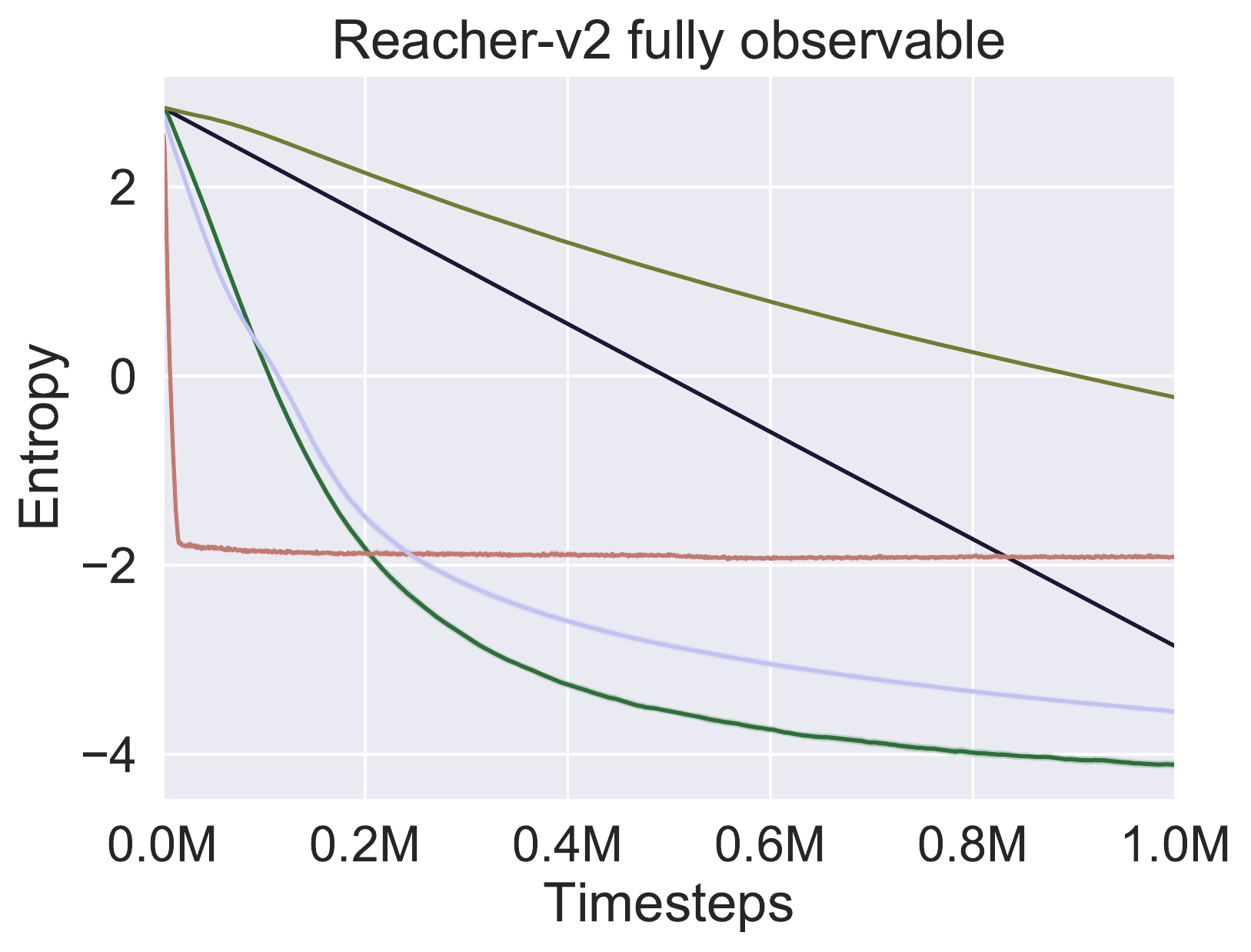}
			\end{subfigure}
			\hfill
			\begin{subfigure}[b]{0.32\textwidth}
				\centering
				\includegraphics[width=\textwidth]{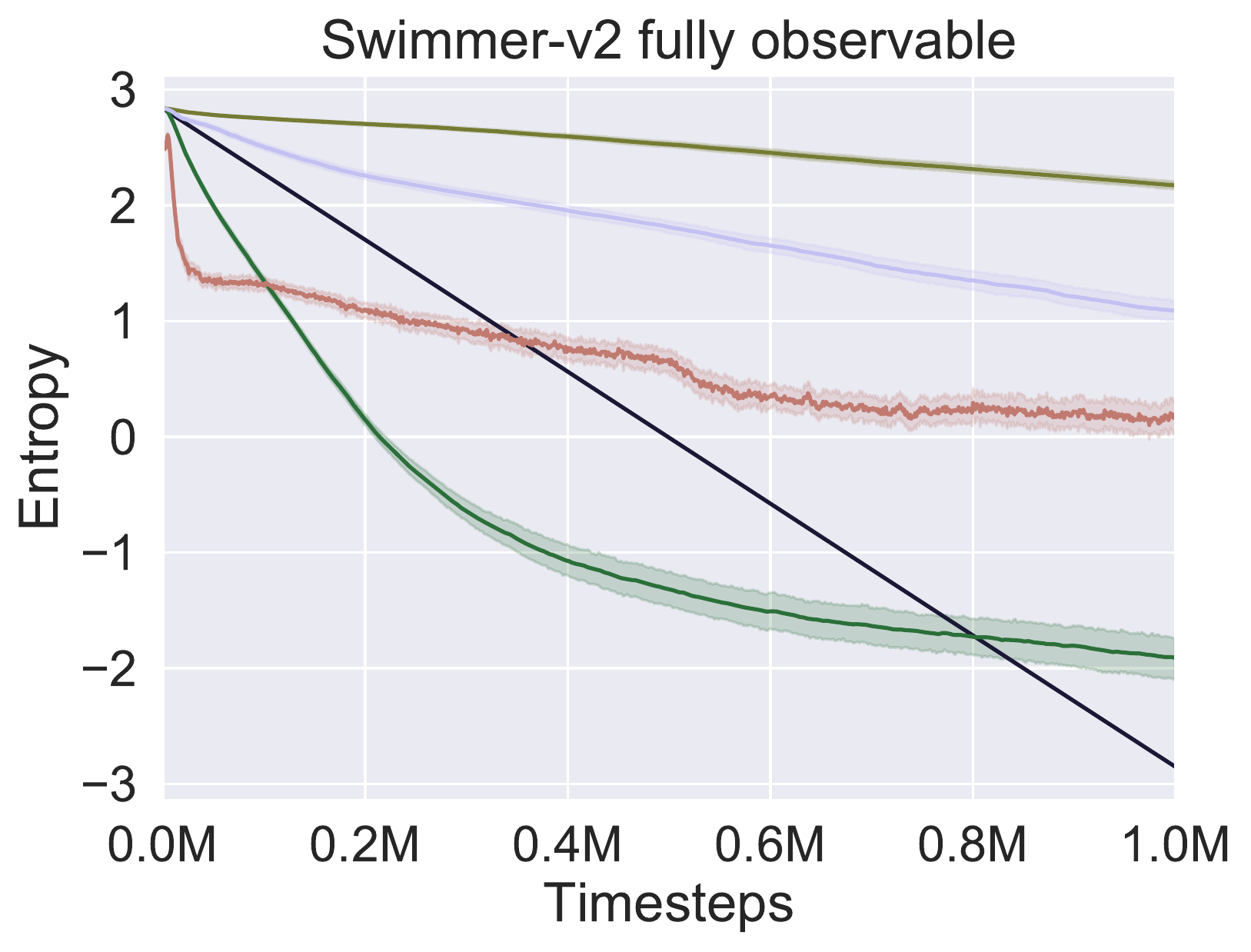}
			\end{subfigure}
			\hfill
			\begin{subfigure}[b]{0.32\textwidth}
				\centering
				\includegraphics[width=\textwidth]{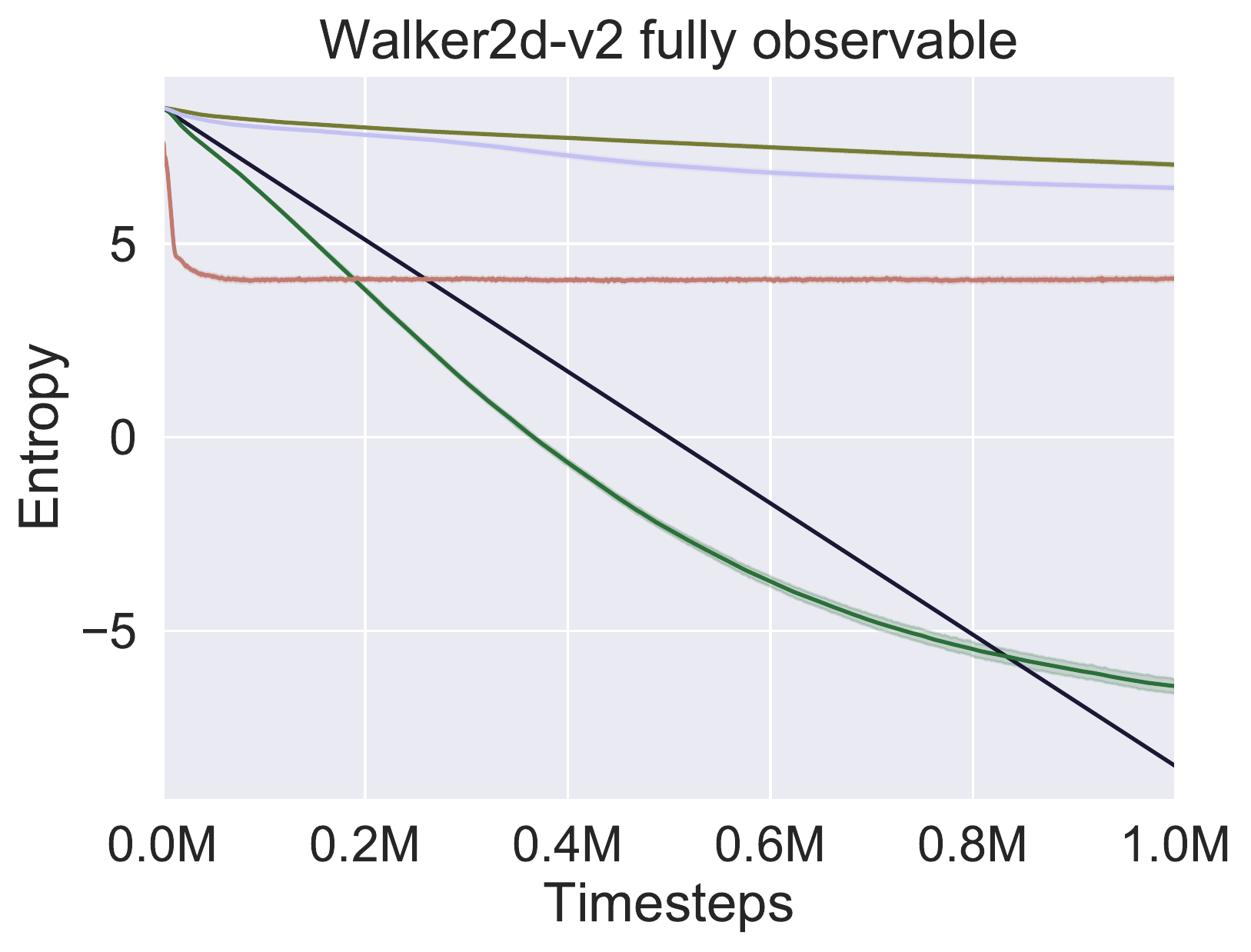}
			\end{subfigure}
			\bigskip
			\begin{subfigure}[b]{0.32\textwidth}
				\centering
				\includegraphics[width=\textwidth]{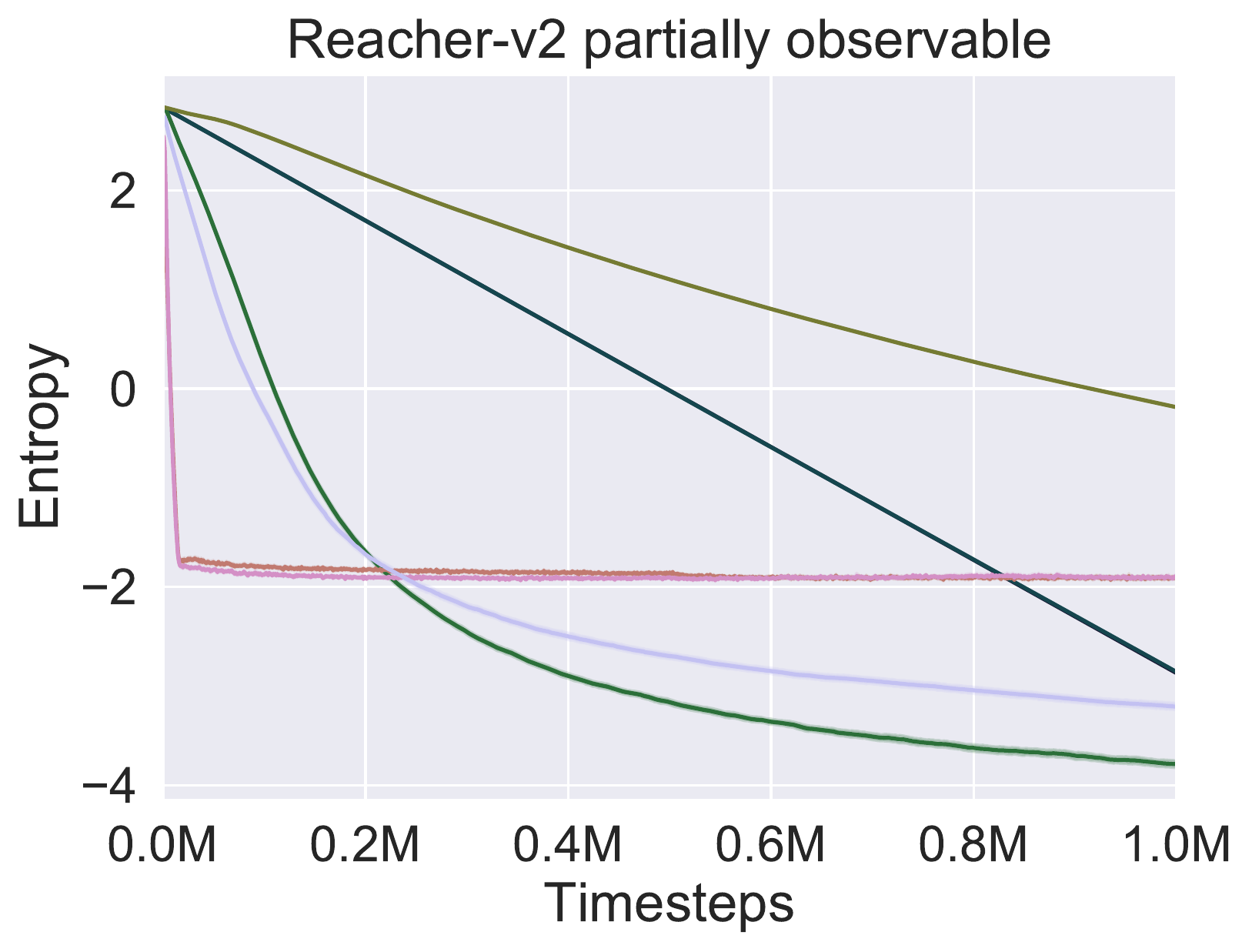}
			\end{subfigure}
			\hfill
			\begin{subfigure}[b]{0.32\textwidth}
				\centering
				\includegraphics[width=\textwidth]{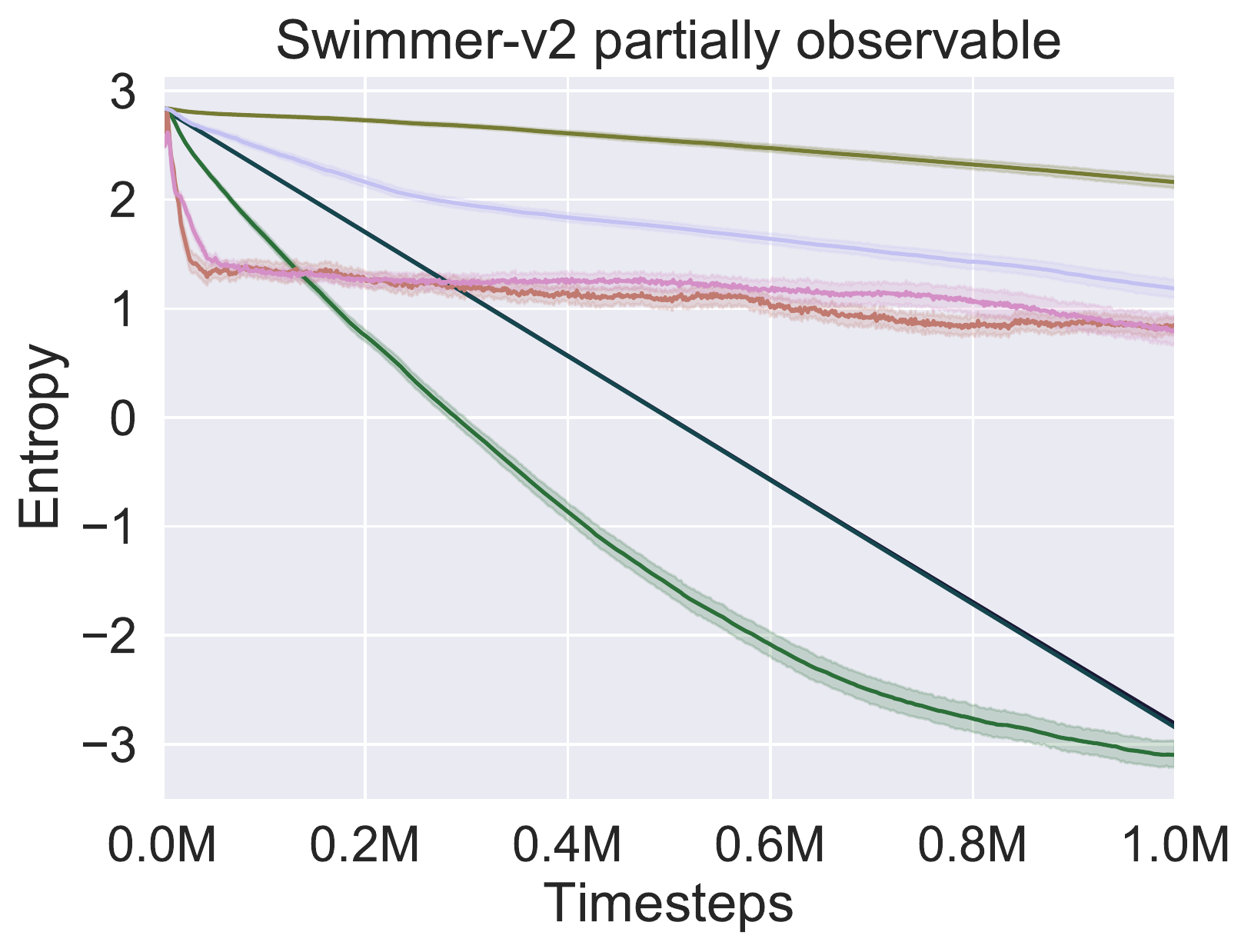}
			\end{subfigure}
			\hfill
			\begin{subfigure}[b]{0.32\textwidth}
				\centering
				\includegraphics[width=\textwidth]{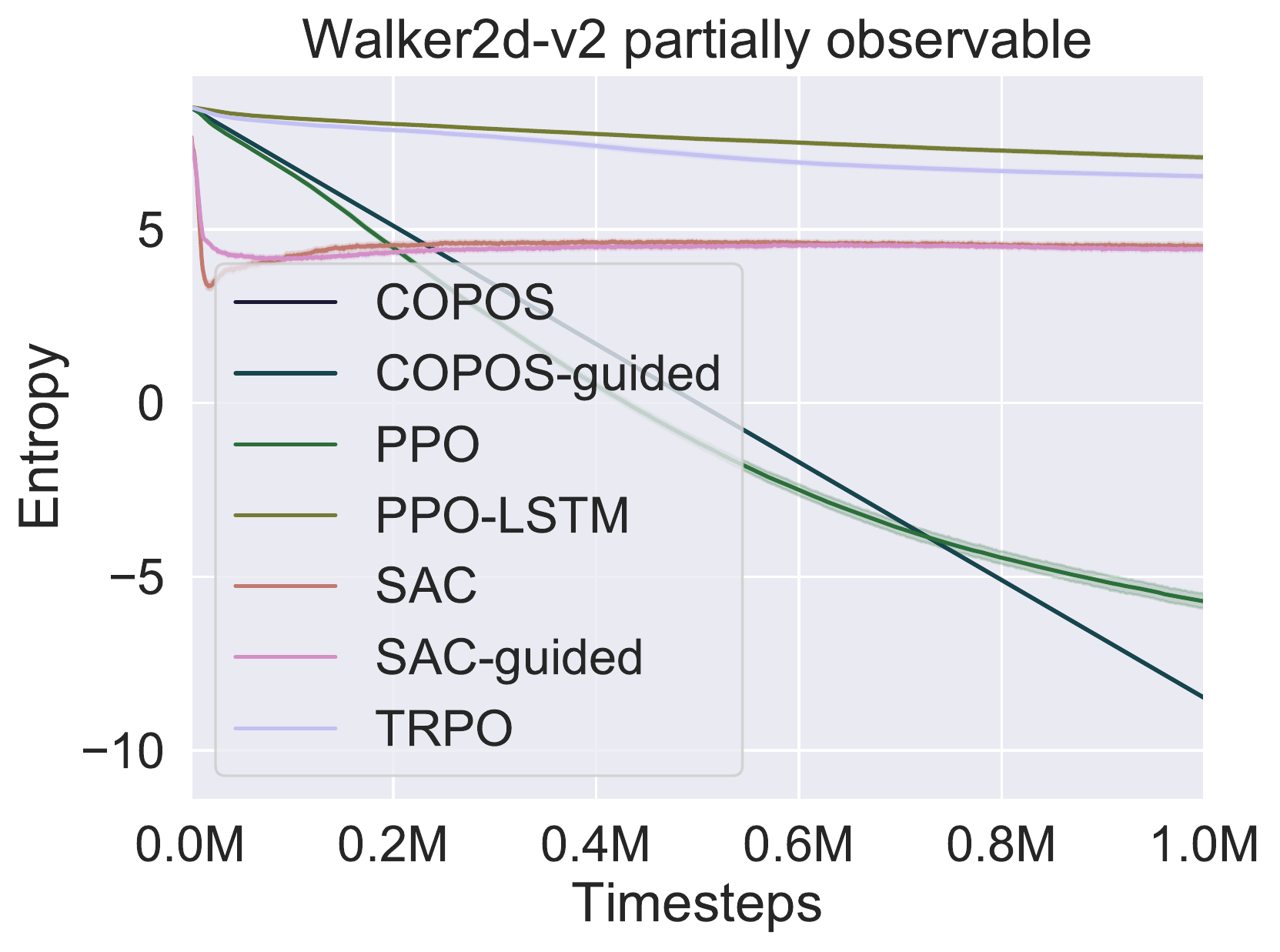}
			\end{subfigure}
			\caption{Entropy for MuJoCo tasks over 50 random seeds. Algorithms were executed for 1 million time steps. Shaded area denotes the bootstrapped 95\% confidence interval.}
			\label{fig:plot-mujoco-entropy}
\end{figure*}

\begin{figure}[ht]
			\centering
			\begin{subfigure}[b]{0.47\columnwidth}
				\centering
				\includegraphics[width=\columnwidth]{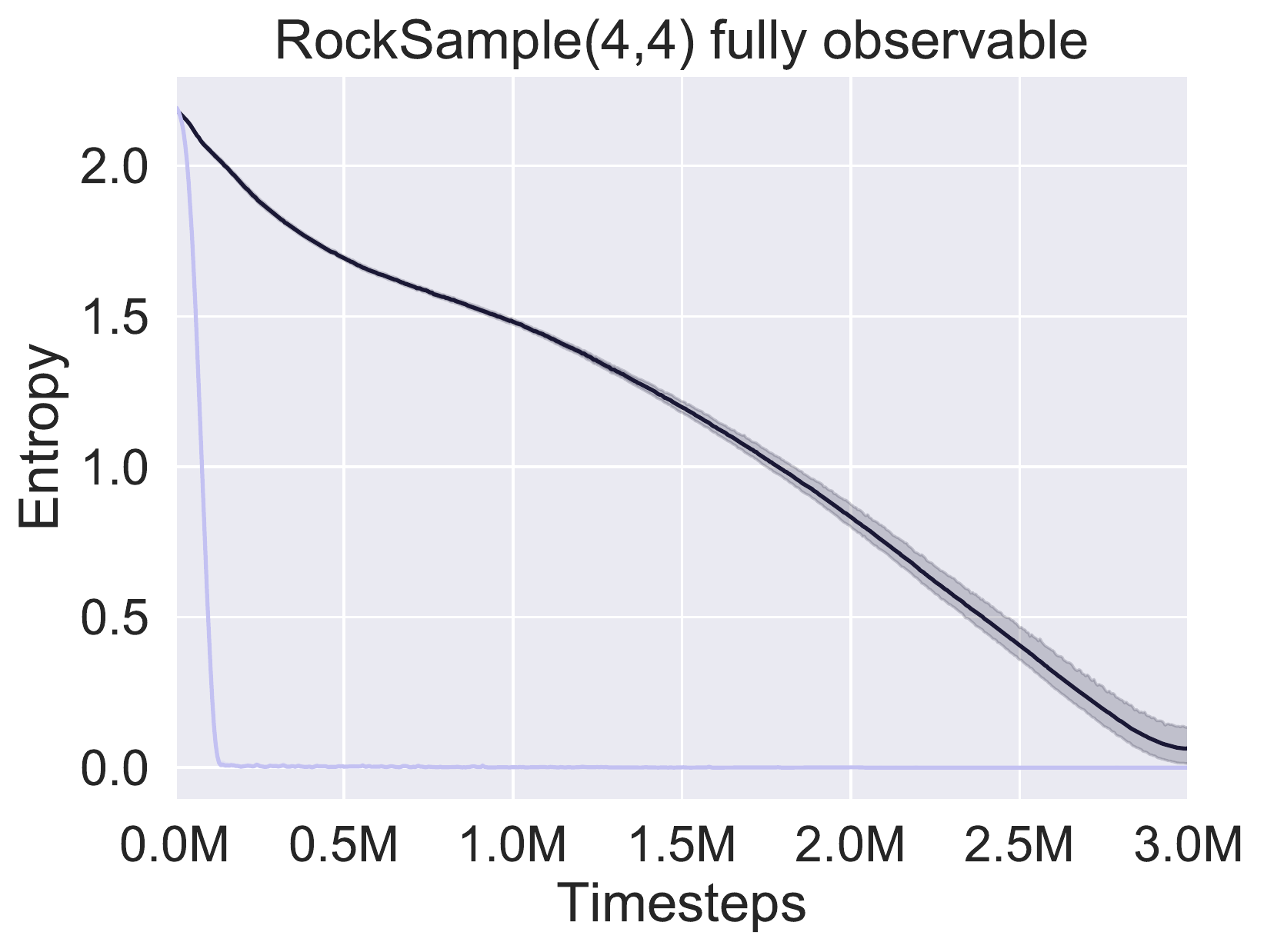}
			\end{subfigure}
			\hfill
			\begin{subfigure}[b]{0.47\columnwidth}
				\centering
				\includegraphics[width=\columnwidth]{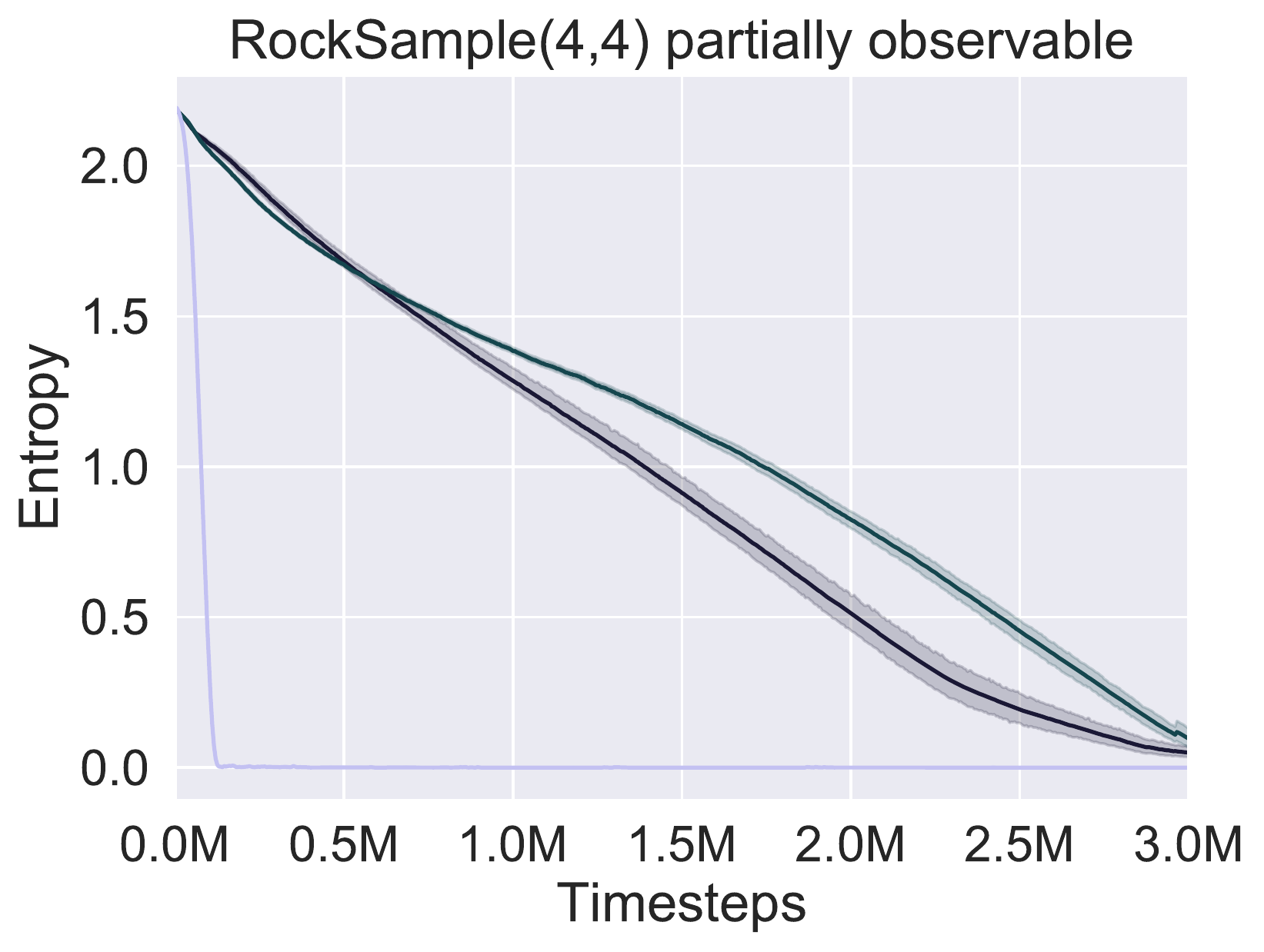}
			\end{subfigure}
			\bigskip
			\begin{subfigure}[b]{0.47\columnwidth}
				\centering
				\includegraphics[width=\columnwidth]{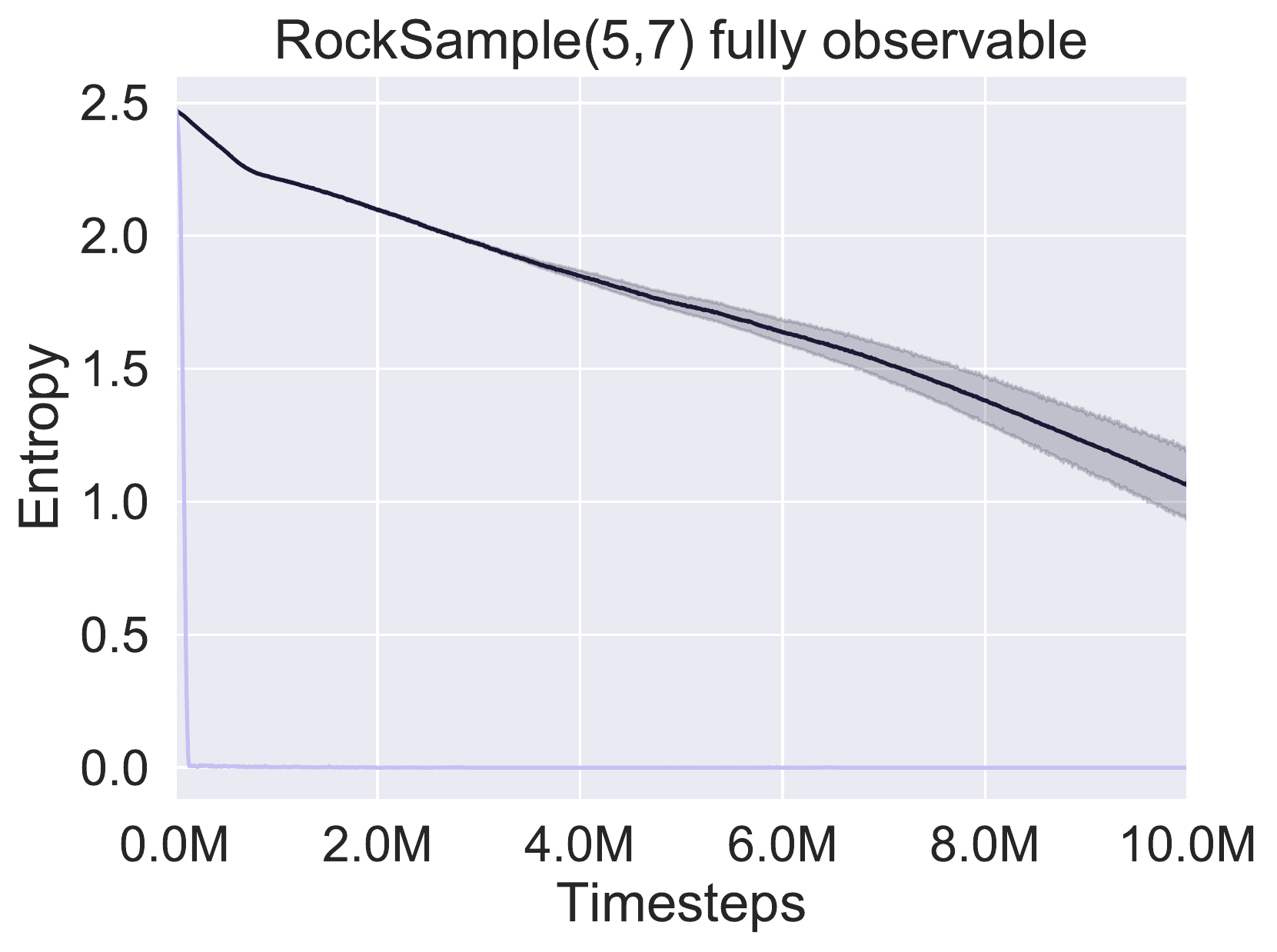}
			\end{subfigure}
			\hfill
			\begin{subfigure}[b]{0.47\columnwidth}
				\centering
				\includegraphics[width=\columnwidth]{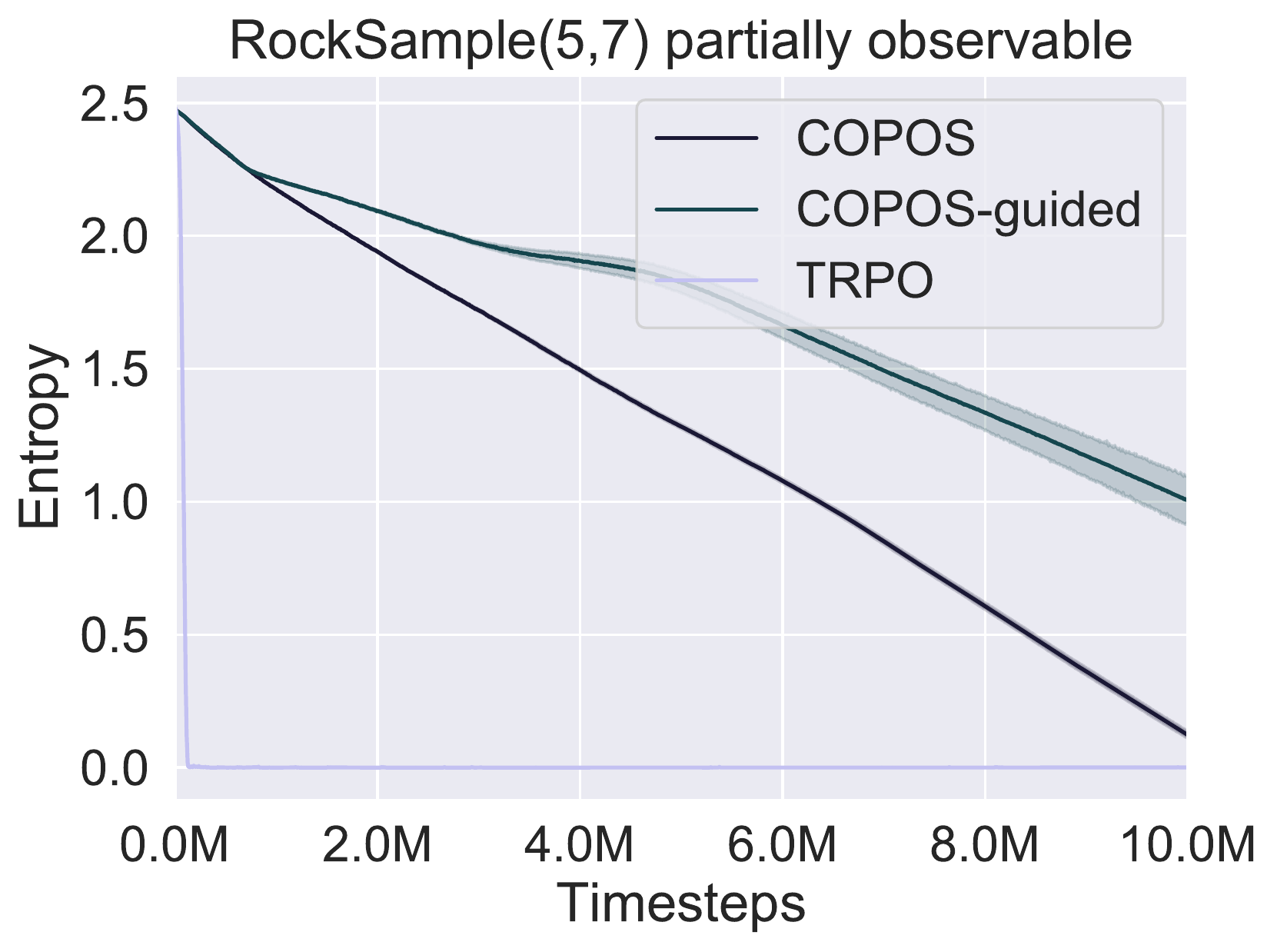}
			\end{subfigure}
			\caption{Entropy for two instances of RockSample, both with full observations (left) and partial observations (right) over 50 random seeds. Algorithms were executed for 5 million time steps on RockSample(4,4) (top) and 10 million time steps on RockSample(5,7) (bottom). Shaded area denotes the bootstrapped 95\% confidence interval.}
			\label{fig:plot-rocksample-entropy}
\end{figure}

Figure \ref{fig:plot-rocksample-return} shows performance plots for two instances of the RockSample problem and Figure \ref{fig:plot-noisylunarlander} shows the performance and entropy plots for the Noisy-LunarLander task. Additional entropy plots can be found in Figure \ref{fig:plot-mujoco-entropy} and Figure \ref{fig:plot-rocksample-entropy}.

\section{MuJoCo POMDP tasks}
See Table \ref{tab:mujoco-ob-ac-space} for a detailed overview of MuJoCo tasks that we modified into partially observable tasks by deactivation some dimensions of the observations vector.

\begin{table}[hb]
			\centering
			\caption{Observation (Ob.) space dimensionality and the dimensions we deactivated for the partially observable versions of the MuJoCo tasks. The column "Description" is for the inactive observation dimensions in the \acs{POMDP} versions.}
			\begin{tabular}{|lccl|}
			\hline
			\textbf{Task} & \textbf{Ob. Space} & \textbf{Inactive Ob.} & \textbf{Description}  \\
			 & \textbf{Dim.} & \textbf{Dim.} & \\
			\hline
			\hline
			HalfCheetah-v2 & 17 & 3-10 & angle of front/back thigh, shin and foot joint\\
							 & & & x and y velocity \\
			Hopper-v2 & 11 & 3-7 & angle of thigh, leg and foot joint \\
							 & & & x and y velocity \\
			InvertedDoublePend.-v2 & 11 & 1-4 & angle of both pole joints\\
							 & & & y position of both pole joints \\
			Reacher-v2 & 11 & 5-11 & target x and y position \\
							 & & & angular velocity of both joints \\
							 & & & distance between fingertip and target \\
			Swimmer-v2 & 8 & 2-6 & angle of both joints, x and y velocity\\
							 & & & angular velocity of torso \\
			Walker2d-v2 & 17 & 9-14 & x and y velocity, angular velocity of torso \\
							 & & & angular velocity of right thigh, shin and foot joint\\
			\hline
			\end{tabular}
			\label{tab:mujoco-ob-ac-space}
\end{table}

\pagebreak
\section{Experiment Parameters}
Additionally to the parameters already described in section 5.1, see Table \ref{tab:params-copos} for \gls{COPOS}, Table \ref{tab:params-ppo} for \gls{PPO}, Table \ref{tab:params-ppo-lstm} for PPO-LSTM, Table \ref{tab:params-sac} for \gls{SAC} and Table \ref{tab:params-trpo} for \gls{TRPO}.

\begin{table}[hb]
	\centering
	\caption{Hyperparameters for running the \acs{COPOS} algorithm.}
	\begin{tabular}{|lc|}
		\hline
		\textbf{Parameter} & \textbf{Value} \\
		\hline
		\hline
		Number of hidden neural network layers & 2 \\
		Units per hidden layer & 32 \\
		Hidden activation function & tanh \\
		Time steps per batch & 5000 (LunarLander and RockSample) or 2000 (MuJoCo) \\
		Maximum KL-divergence ($\beta$) & 0.01 \\
		Maximum difference in entropy ($\epsilon$) & auto (continuous tasks) or 0.01 (discrete tasks) \\
		Conjugate gradient iterations & 10 \\
		Conjugate gradient damping & 0.1 \\
		GAE parameters ($\gamma$ and $\lambda$) & 0.99 and 0.98 \\
		Value function iterations & 5 \\
		Value function step size & 0.001\\
		\hline
	\end{tabular}
	\label{tab:params-copos}
\end{table}

\begin{table}[hb]
	\centering
	\caption{Hyperparameters for running the \acs{PPO} algorithm.}
	\begin{tabular}{|lc|}
		\hline
		\textbf{Parameter} & \textbf{Value} \\
		\hline
		\hline
		Number of hidden neural network layers & 2 \\
		Units per hidden layer & 64 \\
		Hidden activation function & tanh \\
		Time steps per batch & 2048 \\
		Number of mini-batches & 32 \\
		Clip range & 0.2 \\
		GAE parameters ($\gamma$ and $\lambda$) & 0.99 and 0.95 \\
		Number of epochs & 10 \\
		Learning rate & $3\cdot 10^{-4}$ \\
		\hline
	\end{tabular}
	\label{tab:params-ppo}
\end{table}

\begin{table}[hb]
	\centering
	\caption{Hyperparameters for running the PPO-LSTM algorithm.}
	\begin{tabular}{|lc|}
		\hline
		\textbf{Parameter} & \textbf{Value} \\
		\hline
		\hline
		LSTM network size & 32 \\
		Activation function & tanh \\
		Time steps per batch & 2048 \\
		Number of mini-batches & 1 \\
		Clip range & 0.2 \\
		GAE parameters ($\gamma$ and $\lambda$) & 0.99 and 0.95 \\
		Number of epochs & 10 \\
		Learning rate & $3\cdot 10^{-4}$ \\
		\hline
	\end{tabular}
	\label{tab:params-ppo-lstm}
\end{table}

\begin{table}[hb]
	\centering
	\caption{Hyperparameters for running the \acs{SAC} algorithm.}
	\begin{tabular}{|lc|}
		\hline
		\textbf{Parameter} & \textbf{Value} \\
		\hline
		\hline
		Number of hidden neural network layers & 2 \\
		Units per hidden layer & 256 \\
		Hidden activation function & ReLU \\
		Batchsize & 256 \\
		Target update interval & 1 \\
		Gradient steps & 1 \\
		Buffer size & $5 \cdot 10^5$ \\
		Learning rate & $3\cdot 10^{-4}$ \\
		\hline
	\end{tabular}
	\label{tab:params-sac}
\end{table}

\begin{table}[hb]
	\centering
	\caption{Hyperparameters for running the \acs{TRPO} algorithm.}
	\begin{tabular}{|lc|}
		\hline
		\textbf{Parameter} & \textbf{Value} \\
		\hline
		\hline
		Number of hidden neural network layers & 2 \\
		Units per hidden layer & 32 \\
		Hidden activation function & tanh \\
		Time steps per batch & 5000 (LunarLander and RockSample) or 2000 (MuJoCo) \\
		Maximum KL-divergence & 0.01 \\
		Conjugate gradient iterations & 10 \\
		Conjugate gradient damping & 0.1 \\
		GAE parameters ($\gamma$ and $\lambda$) & 0.99 and 0.98 \\
		Value function iterations & 5 \\
		Value function step size & 0.001\\
		\hline
	\end{tabular}
	\label{tab:params-trpo}
\end{table}


\end{document}